\documentclass[a4paper]{article}

\date{}
\usepackage[utf8]{inputenc}
\usepackage[a4paper, total={6in, 9in}]{geometry}
\usepackage{authblk}
\usepackage[square,numbers]{natbib}
\usepackage{graphicx}
\usepackage{amsmath}
\usepackage{algorithm}
\usepackage{algpseudocode}
\usepackage{amsfonts}
\usepackage{bm}
\usepackage{amssymb}
\usepackage{textcomp}
\usepackage{latexsym}
\usepackage{hyperref}
\usepackage{url}
\usepackage{framed}
\usepackage{comment}
\renewenvironment{framed}[1][\hsize]
   {\MakeFramed{\hsize.95\linewidth\advance\hsize-\width \FrameRestore}}%
   {\endMakeFramed}
\usepackage{csquotes}   
\usepackage{tabularx}
\usepackage{subcaption}
\usepackage{makecell}
\usepackage{multirow}
\usepackage{enumitem}
\usepackage{xcolor}

\newcommand{\vy}{{\bm{y}}}
\newcommand{\vx}{{\bm{x}}}

\newcommand{\vE}{{\bm{E}}}
\newcommand{\vb}{{\bm{b}}}
\newcommand{\vc}{{\bm{c}}}
\newcommand{\vs}{{\bm{s}}}

\newcommand{\vu}{{\bm{u}}}
\newcommand{\vW}{{\bm{W}}}

\newcommand{\vd}{{\bm{d}}}
\newcommand{\vX}{{\bm{X}}}
\newcommand{\vY}{{\bm{Y}}}

\newcommand{\RNN}{\textrm{RNN}}
\newcommand{\softmax}{\textrm{softmax}}
\newcommand{\vtheta}{{\bm{\theta}}}

\newcommand{\local}{{\textrm{local}}}
\newcommand{\mglobal}{{\textrm{global}}}
\newcommand{\src}{{\textrm{src}}}

\begin{document}

\title{A Survey on Document-level Neural Machine Translation: Methods and Evaluation}

\author{Sameen Maruf, Fahimeh Saleh and Gholamreza Haffari\thanks{This work is supported by a Google Faculty Research Award and ARC FT190100039 to G.H. The authors are grateful to the anonymous reviewers for their valuable feedback and to George Foster for his comments on the final version of this paper.}
}
\affil{%
  Faculty of Information Technology, Monash University, \\Clayton VIC, Australia\\
	\texttt{\{firstname.lastname\}$@$monash.edu}}

\maketitle

\begin{abstract}
Machine translation (MT) is an important task in natural language processing (NLP) as it automates the
translation process and reduces the reliance on human translators. 
With the resurgence of neural networks, the translation quality surpasses that of the translations obtained using statistical techniques for most language-pairs. Up until a few years ago, almost all of the neural translation models translated sentences \emph{independently}, without incorporating the wider \emph{document-context} and inter-dependencies among the sentences. \emph{The aim of this survey paper is to highlight the major works that have been undertaken in the space of document-level machine translation after the neural revolution, 
so that researchers can recognise 
the current state and future directions of this field.} 
We provide an organisation of the literature based on novelties in modelling and architectures as well as training and decoding strategies. 
%
%
In addition, we  cover evaluation strategies that have been introduced to account for the improvements in document MT, including automatic metrics and discourse-targeted test sets. We conclude by presenting possible avenues for future exploration in this research field.
\end{abstract}

\section{Introduction}\label{sec:intro}

Machine translation (MT) is the process of automating translation between natural languages with the aid of computers. Translation, in itself, is a difficult task even for humans as it requires a thorough understanding of the source text and a good knowledge of the target language, hence requiring the human translators to have high degree of proficiency in both languages. Due to the dearth of professional translators and the rapid need of availability of multilingual digital content, for example, on the Internet, MT has grown immensely over the past few decades for the purposes of international communication. The global market for machine translation is expected\footnote{\url{https://www.psmarketresearch.com/market-analysis/machine-translation-market}} to reach USD 2.3 billion by 2023, with an expected average annual growth rate of 20.9\%. The global deep learning market size was valued\footnote{\url{http://www.grandviewresearch.com/industry-analysis/deep-learning-market}} at USD 272.0 million in 2016, and is expected\footnote{\url{http://www.grandviewresearch.com/press-release/global-deep-learning-market}} to reach USD 10.2 billion by 2025. Higher accuracy MT can better facilitate interlingual communication in various areas, e.g., business, education, and government. It contributes to the economic growth by reducing the cost of production and trade, and providing better marketing opportunities for businesses.

Up until a few years ago, MT was mostly formalised through statistical techniques, hence very aptly named statistical machine translation (SMT), and involved meticulously crafting features to extract implicit information from corpora of bilingual sentence-pairs \cite{Brown:1993}. These hand-engineered features were an intrinsic part of SMT and were one of the reasons behind its inflexibility. MT has come a long way since then to the state-of-the-art neural machine translation (NMT) systems \cite{Sutskever:14, Bahdanau:15, Vaswani:17}, which are based on neural network architectures requiring little to no feature engineering. 
MT systems have seen rapid improvements in the past few years, and this has added to their popularity among the general public \cite{Metz2016, Kraus2016} and the research community \cite{Wu:2016, Johnson:17, Dehghani:18}.

\emph{Inspite of its success,  MT has been mainly based on strong independence and locality assumptions, that is, translating sentences in isolation independent of their document-level inter-dependencies.}
Text, on the contrary, does not consist of isolated, unrelated elements, but of collocated and structured group of sentences bound together by complex linguistic elements, referred to as the \emph{discourse} \cite{Jurafsky:2009}. Ignoring the inter-relations among these discourse elements results in translations which may seemingly be perceived as good but lack crucial properties of the text. In other words, ignoring the document context in translation hinders the transmission of the intended meaning. 

The use of context in MT  has been advocated by MT pioneers for decades \cite{Barhillel1960, Sennrich:WMT18}. However, most of the works that endeavoured to include document context in SMT were unable to yield significant improvements due to the limitations of SMT and concerns from the MT community about its computational efficiency and tractability. Recently, with the increase in computational power and the wide-scale application of neural networks to machine translation, the community has finally been in a position to forego the independence constraints that have impeded the progress in MT since long. There has been a lot of attention and works focusing on {document-level neural machine translation} (definition to follow) in the past few years. \textit{We believe that it is the right time to take stock of what has been achieved in document NMT,  so that researchers  can get a bigger picture of where this line of research stands.}
\\
\\
\textbf{Illustrating examples.}
To illustrate the need of incorporating discourse in MT and highlight the limitations of sentence-based translation, let us give a Chinese$\rightarrow$English example translation \cite{Sennrich:WMT18}:
\begin{framed}
\setlength{\parindent}{2pt}
\begin{displayquote}
Members of the public who find their cars obstructed by unfamiliar vehicles during their daily journeys can use the ``Twitter Move Car'' feature to address this distress when the driver of the unfamiliar vehicle cannot be reached.
\end{displayquote}
\end{framed}
\noindent At first glance, it is difficult to distinguish this MT output from a human translation. However, let us now provide a translation of the complete text as generated by an MT system:

\begin{framed}
\setlength{\parindent}{2pt}
\begin{displayquote}
Members of the public who find their cars obstructed by unfamiliar vehicles during their daily journeys can use the {\textcolor{red}{``Twitter Move Car''}} feature to address this distress when the driver of the unfamiliar vehicle cannot be reached. On August 11, Xi'an traffic police WeChat service number ``Xi'an traffic police'' launched {\textcolor{red}{``WeChat mobile''}} service. With the launch of the service, members of the public can tackle such problems in their daily lives by using the {\textcolor{red}{``WeChat Move''}} feature when an unfamiliar vehicle obstructs the movement of their vehicle while the driver is not at the scene. [$\hdots$]
\end{displayquote}
\end{framed}

\noindent An obvious problem with the generated text is the inconsistent translation of the name of the service ``WeChat Move the Car''. In other words, although it seems that the sentence-based translation is adequate on its own, it contains some ambiguous words which are inconsistent with the rest of the text. 
Let us look at another example translation for Urdu$\rightarrow$English generated by Google Translate:
\begin{framed}
\setlength{\parindent}{2pt}
\begin{displayquote}
My grandfather’s legs have failed because of the {\textcolor{blue}{fluid}}. {\textcolor{red}{He}} had another {\textcolor{blue}{visit}} today. Then {\textcolor{red}{his}} {\textcolor{blue}{nature}} worsened. {\textcolor{red}{They}} can not speak for a few moments.
\end{displayquote}
\end{framed}
\noindent Even without having access to the Urdu source, we can concur that the English target text has some prominent issues, including inconsistent usage of pronouns ({\it he}, {\it they}) and ambiguous words ({\it fluid}, {\it visit}, {\it nature}), resulting in non-fluency and miscomprehension of the target text. 
From the previous two examples, it should be clear that despite the success of MT, it will never achieve human-level translation if it continues to be grounded on sentence-independence or locality assumptions. \\
\\
\textbf{Structure and scope of this survey.}
This survey aims to highlight the major works that have been undertaken in the space of \textit{document-level neural machine translation} (Section
~\ref{sec:disnmt}).\footnote{We do not cover the parallel works that incorporate discourse in SMT in this survey but would like to direct interested readers to the literature reviews in \cite{Hardmeier:2014, Smith:2018} that cover this topic in detail.} By document-level MT, we mean works which utilise inter-sentential context information comprising discourse aspects of a document or surrounding sentences in the document. In addition to this, we also cover the evaluation strategies that have been introduced to account for improvements in this domain (Section~\ref{sec:diseval}). We conclude by presenting avenues for future research (Section~\ref{sec:conc}). 

Before moving on with the main agenda, we briefly describe the basics of sentence-level neural MT models and their evaluation in the following section. We do not endeavour to cover the entire breadth of work in sentence-level NMT (as that would require an independent survey) but just enough detail to establish the foundations of document-level NMT necessary for understanding the progress in this field. For a more comprehensive review of sentence-level NMT (and SMT), we direct the interested readers to \cite{Stahlberg:review} (and \cite{Lopez:2008}).

\section{Background: Sentence-level Neural Machine Translation}\label{sec:nmt}

With the resurgence of neural-based approaches and their application to machine translation (since 2014), the field of neural machine translation has grown leaps and bounds starting a new era in MT for both research and commercial purposes. The main advantage that NMT has over its predecessors is having a single end-to-end model whose parameters can be jointly optimised with respect to a training objective.


Mathematically, the goal of sentence-level NMT is to find the most probable target sequence $\hat{\vy}$ given a source sentence, that is:
\begin{equation}\label{eq:inference}
\hat{\vy}=\underset{\vy}{\text{arg max }}P(\vy\mid {\vx})
\end{equation}

The conditional probability $P(\vy\mid{\vx})$ is modelled using neural networks, where $\vx = (x_1,\hdots,x_M)$ is the input (source) sequence and $\vy = (y_1,\hdots,y_N)$ is the output (target) sequence. 
The conditional probability of a target sentence $\vy$ given the source sentence $\vx$ is decomposed as:
\begin{equation}\label{eq:cond}
P_{\vtheta}(\vy\mid\vx)=\prod_{n=1}^N P_{\vtheta}(y_n\mid\vy_{<n}, \vx)
\end{equation}
where $\vtheta$ denotes the learnable parameters of the neural network, $y_n$ is the current target word and $\vy_{<n}$ are the previously generated words. 

\begin{figure}[t]
 \centering
  \includegraphics[width=0.8\linewidth]{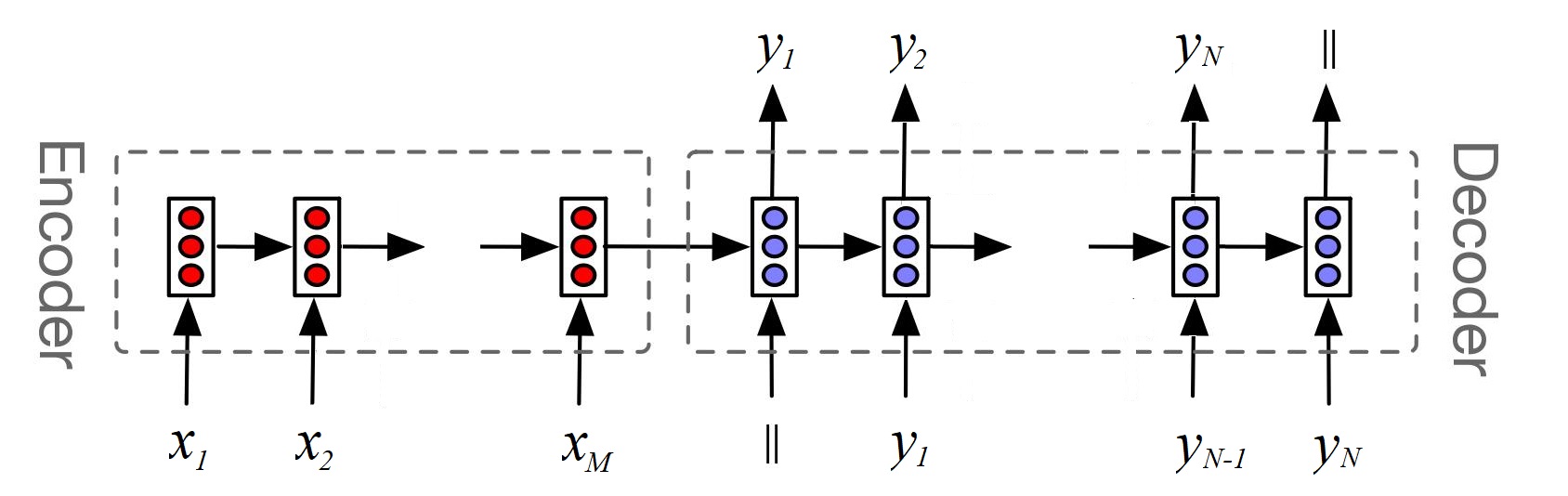}
  \caption{A general overview of an encoder-decoder model for sentence-level NMT.}
 \label{fig:genmt}
\end{figure}

\subsection{Neural encoder-decoder architectures}
\begin{figure}[t]
\centering
\begin{subfigure}{.45\textwidth}
  \centering
  \includegraphics[width=.7\linewidth]{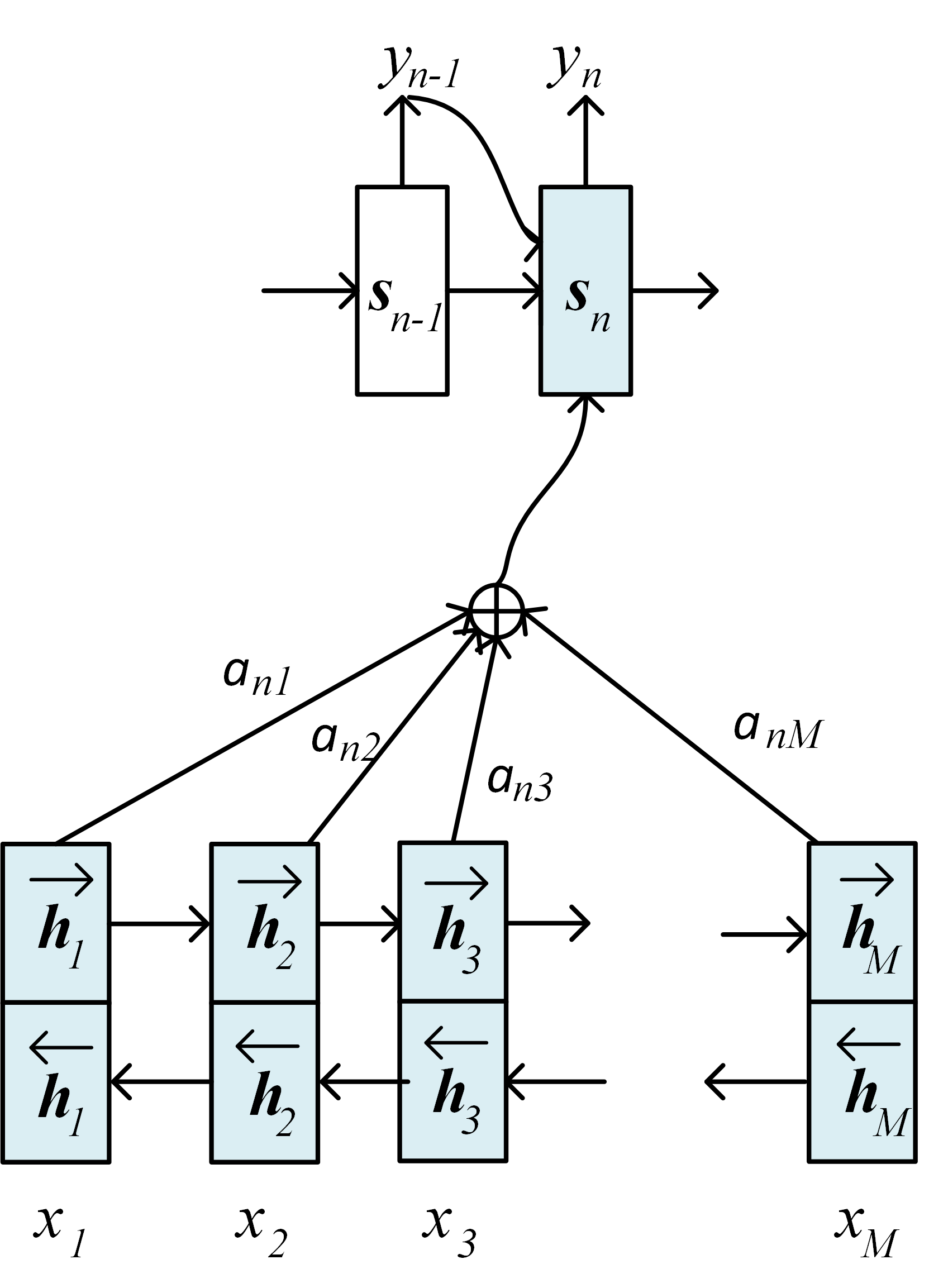}
  \caption{Attentional RNN-based architecture \cite{Bahdanau:15}}
  \label{fig:rnnnmt}
\end{subfigure}%
\begin{subfigure}{.55\textwidth}
  \centering
  \includegraphics[width=.75\linewidth]{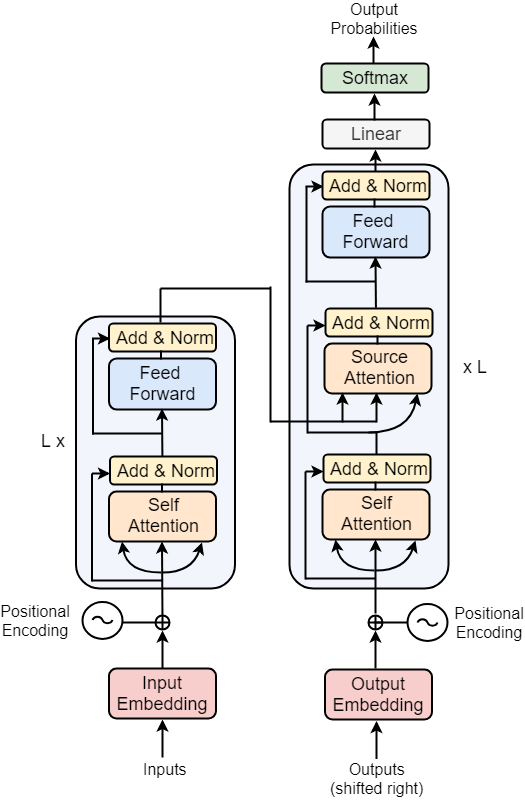}
  \caption{Transformer architecture \cite{Vaswani:17}}
  \label{fig:transformer}
\end{subfigure}
\caption{Neural encoder-decoder MT architectures relevant to this survey.}
\vspace{-1mm}
\label{fig:test}
\end{figure}

NMT models, in general, are based on an encoder-decoder framework (Figure~\ref{fig:genmt}), where the encoder reads the source sentence to compute a real-valued representation and the decoder generates the target translation one word at a time given the previously computed source representation. The initial model by Sutskever et al. \citep{Sutskever:14} used a fixed representation of the source sentence to generate the target sentence. It was quickly replaced by the attention-based encoder-decoder architecture of Bahdanau et al. (Figure~\ref{fig:rnnnmt}) which generates a dynamic context representation \citep{Bahdanau:15}. These models were based on recurrent neural networks (RNNs) \cite{Elman:90}, which use recurrent connections to exhibit temporal dynamic behavior over time, and were thus considerably suitable for modelling sequential information. However, the major drawback of such sequential computation was that it hindered parallelisation within training examples and became a bottleneck when processing long sentences. Most recently, a new model architecture, the Transformer \cite{Vaswani:17}, was introduced which is based solely on attention mechanisms, dispensing with the recurrence entirely (Figure~\ref{fig:transformer}). It has proved to achieve state-of-the-art results on several language-pairs.

There have been numerous other NMT models such as those based on convolutional neural networks (CNNs) \cite{Gehring:2017, Gehring:17} and variations of the Transformer such as the Universal Transformer \cite{Dehghani:18}. However, in the rest of this section, we will only cover the attentional RNN model and the Transformer as these two have been extensively used in the field of document-level NMT (all document-level NMT models are grounded on either one of them). Further, a basic knowledge of these two models is necessary to develop a thorough understanding of Section~\ref{sec:disnmt}. \\
\\
\textbf{RNN-based Architecture.} Bahdanau et al.'s \textit{encoder} is a bi-directional (forward and backward) RNN \cite{Elman:90} whose hidden states represent individual words of the source sentence. The forward and backward RNNs run over the source sentence in a left-to-right and right-to-left direction 
and each word in the source sentence is then represented by the concatenation of the corresponding bidirectional hidden states (see bottom of Figure~\ref{fig:rnnnmt}). 
These representations capture information not only of the corresponding word but also of other words in the sentence, i.e., the sentential context. 

An integral component of the RNN-based NMT architecture is the attention mechanism. This enables the decoder to dynamically \textit{attend} to relevant parts of the source sentence at each step of generating the target sentence. The dynamic context vector $\vc_n$ (also referred to as the attentional vector) is computed as a weighted linear combination of the hidden states produced by the bi-directional RNNs in the encoder, 
where the weights ($\alpha$'s in Figure~\ref{fig:rnnnmt}) can be thought of as the \textit{alignment probability} between a target symbol at position $n$ and a source symbol at position $m$. 

The final component of an encoder-decoder architecture for NMT is the \textit{decoder}. The backbone of the decoder is a uni-directional RNN which generates words of the target translation one-by-one in a left-to-right fashion. The decoder hidden state is computed as follows:
\begin{eqnarray}\label{eq:nmtdecoder}
\vs_n&=& \RNN(\vs_{n-1},\vE_{\bm{T}}[y_{n-1}],\vc_n)
\end{eqnarray}
where $\vs_{n-1}$ is the previous decoder state, $\vE_{\bm{T}}[y_n]$ is embedding of the word $y_n$ from the embedding table $\vE_{\bm{T}}$ of the target language, and $\vc_n$ is the dynamic context vector 
defined previously. 
The probability of generation of each word $y_n$ is then conditioned on all of the previously generated words $\vy_{<n}$ via the state of the RNN decoder $\vs_n$, and the source sentence via $\vc_n$:
\begin{eqnarray}
\vu_n&=& \tanh(\vs_n + \vW_{\bm{uc}}\vc_n +\vW_{\bm{un}}\vE_{\bm{T}}[y_{n-1}]) \nonumber \\
P_{\vtheta}(y_n\mid\vy_{<n}, \vx)&=& \softmax(\vW_{\vy}\vu_{n}+\vb_{\vy})\\
y_n &\sim& P_{\vtheta}(y_n\mid\vy_{<n}, \vx)
\end{eqnarray}
where $\vW$ matrices and $\vb_{\vy}$ vector are also parameters of the NMT model and the input to the softmax (linear transformation of $\vu_n$) is a score vector over the target vocabulary. 
\\
\\
\textbf{Transformer-based Architecture.} RNN-based NMT models have two major limitations. The first limitation is the sequential nature of RNNs, that is for processing each input token, the model has to wait until all previous input tokens have been processed, which proves to be a bottleneck when processing long sequences. The second limitation is learning long-range dependencies among the tokens within a sequence. The number of operations required to relate signals from two arbitrary input or output positions grows with the distance between positions, making it difficult to learn complex dependencies between distant positions. The recent \textit{Transformer} architecture \cite{Vaswani:17} circumvents these limitations by having a model that is still based on the philosophy of encoder-decoder, but instead of employing recurrence, uses stacked self-attention and point-wise, fully connected layers for both the encoder and decoder (Figure~\ref{fig:transformer}).

The \textit{encoder stack} is composed of $L$ identical layers, each containing two sub-layers. The first sub-layer comprises multi-head self-attention 
allowing each position in the encoder to attend to all positions in the previous layer of the encoder, while the second sub-layer is a feed-forward network which uses two linear transformations with a ReLU activation function. 
The \textit{decoder stack} is also composed of $L$ identical layers. In addition to the two sub-layers previously described, it also comprises a third sub-layer, which performs multi-head attention over the output of the corresponding encoder layer. Masking is used in the self-attention sub-layer in the decoder stack 
to prevent positions from attending to subsequent positions and avoid leftward flow of information.

\subsection{Training}
All parameters in the encoder-decoder architectures (RNN-based or Transformer) are jointly trained via backpropagation \citep{Lecun:1988, Rumelhart:1986} to minimise the negative log-likelihood (conditional) over the training set. The conditional log-likelihood is defined as the sum of the log-probability of predicting a correct symbol $y_n$ in the output sequence for each instance $\vx$ in the training set $\mathcal{D}$. Thus, we want to find the optimum set of parameters ${\vtheta}^*$ as follows:
\begin{eqnarray}\label{eq:learning}
{\vtheta}^*&=&\underset{\vtheta}{\text{arg min }}\sum_{(\vx,\vy)\in \mathcal{D}}{-}\log P_{\vtheta}(\vy\mid {\vx})\\
&=&\underset{\vtheta}{\text{arg min }}\sum_{(\vx,\vy)\in \mathcal{D}}\sum_{n=1}^{|\vy|}{-}\log P_{\vtheta}(y_n\mid \vy_{<n},{\vx})
\end{eqnarray}

\subsection{Decoding}
Having trained an NMT model, we need to be able to use it to translate or decode unseen source sentences. The best output sequence for a given input sequence is produced by: 
\begin{eqnarray}
\hat{\vy} = {\underset{\vy}{\text{arg max }}}{P_{\vtheta}(\vy|\vx)}
\end{eqnarray}
Solving this optimisation problem exactly is computationally hard, and hence an approximate solution is obtained using greedy decoding or beam search. 

The basic idea of greedy decoding is to pick the most likely word (having the highest probability) at each decoding step until the end-of-sentence token is generated. Beam search \cite{Graves:12}, on the other hand, keeps a fixed number (\textit{b}) of translation hypotheses with the highest log-probability at each timestep. 
A complete hypothesis (containing the end-of-sentence token) is added to the final candidate list. The algorithm then picks the translation with the highest log-probability (normalised by the number of target words) from this list. If the number of candidates at each timestep is chosen to be one, beam search reduces to greedy decoding. In practice, the translation quality obtained via beam search (size of 4) is significantly better than that obtained via greedy decoding. However, beam search is computationally very expensive (25\%-50\% slower depending on the base architecture and the beam size) in comparison to greedy decoding \citep{Chen:2018}.

\subsection{Evaluation}
To evaluate the quality of the generated translations, numerous automatic evaluation metrics have been proposed. Here we mention two of the most popular n-gram matching ones: BLEU and METEOR, as only these are relevant for the purposes of this survey.\footnote{We direct the interested readers to \cite{Chatzikoumi:20} for an in-depth review of MT evaluation and to \cite{Zhang2020BERT} for an empirical comparison.} 

BLEU (Bilingual Evaluation Understudy) \citep{Papineni:02} has been a de-facto standard for evaluating translation outputs since it was first proposed in 2002. The core idea is to aggregate the count of words and phrases (n-grams) that
overlap between machine and reference translations. 
The BLEU metric ranges from 0 to 1, where 1 means an identical output with the reference. 

Although BLEU correlates well with human judgment \citep{Papineni:02}, it relies on precision alone and does not take into account recall -- the proportion of the  matched n-grams out of the total number of n-grams in the reference translation. METEOR \citep{Banerjee:05, Lavie:07} was proposed to address the shortcomings of BLEU. It scores a translation output by performing a word-to-word alignment between the translation output and a given reference translation. 
The alignments are produced via a sequence of word-mapping modules, that is, if the words are exactly the same, same after they are stemmed using the Porter stemmer, and if they are synonyms of each other. 
%
After obtaining the final alignment, METEOR computes $F_{mean}$, which is just the parameterised harmonic mean of unigram precision and recall \citep{Rijsbergen:1979}. 
METEOR has also demonstrated to have a high level of correlation with human judgment, even outperforming that of BLEU \citep{Banerjee:05}. 

To make the results of the aforementioned MT evaluation metrics more reliable, a statistical significance test should be performed \citep{Koehn:04} which indicates whether the difference in translation quality of two or more systems is due to a difference in true system quality. 
\section{Document-level Neural Machine Translation}\label{sec:disnmt}

Most MT models are built on strong independence and locality assumptions, either intra-sentential as done by phrase-based models or inter-sentential as done by even the most advanced NMT models today. From a linguistics perspective, this assumption is invalid, as any piece of text is much more than just a single sentence and making this assumption means ignoring the underlying discourse structure of the text and still hoping that the translation would not fall short. 

\begin{figure}[t]
 \centering
  \includegraphics[width=0.9\linewidth]{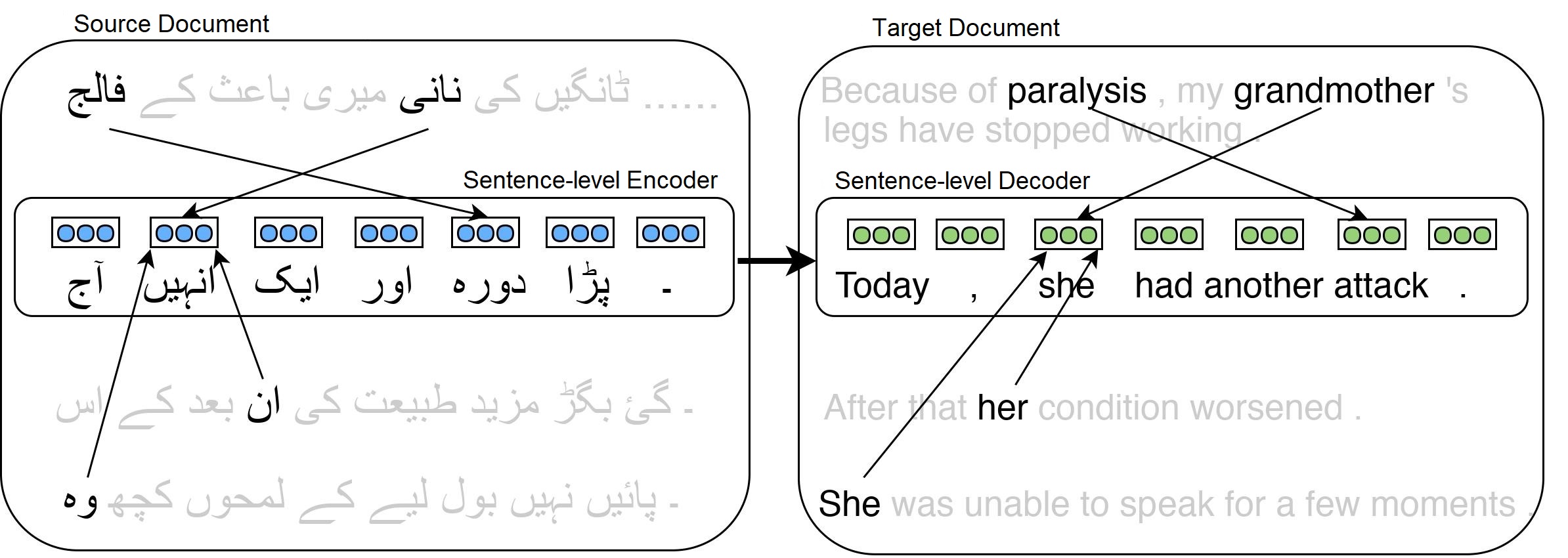}
  \caption{A generic document-level NMT model; the target-side is the human translation of the last excerpt in Section~\ref{sec:intro}.}
 \label{fig:doc-discourse}
 \vspace{-3mm}
\end{figure}

\textit{Discourse} is defined as a group of sentences that are contiguous, structured and exhibit coherency \citep{Jurafsky:2009}. To elaborate why discourse is essential for MT, let us provide a concrete example with the help of Figure~\ref{fig:doc-discourse}. Urdu pronouns in the second sentence are gender-neutral and to accurately translate them to English, one needs access to the noun (the Urdu source word for \textit{grandmother} present in the previous sentence or its English translation) or another target pronoun (\textit{her} and \textit{she} present in the following target sentences).\footnote{It should be noted that in this example, the Urdu source pronouns for \textit{she} in the second and fourth sentence are different.} In addition, the Urdu source for the word \textit{attack} is ambiguous and unless one has access to the Urdu source for the target word \textit{paralysis}, it cannot be accurately translated. Thus, the only way an MT model can accurately translate the second sentence is if it has access to the previous and following sentences in the document. Table~\ref{table:def} provides definitions of a few discourse phenomena that have been widely discussed in the document MT literature.

Although the problem of machine translation itself has been around for decades, the works which have tried to address issues that arise from not having access to the document-level discourse are still just brushing the surface with more research yet to be undertaken. For NMT in particular, up until a few years ago, there was rarely any work that tried to incorporate discourse information explicitly or implicitly. But with most sentence-based NMT systems achieving state-of-the-art performance compared to their SMT counterparts, this area of research has finally started to gain the popularity it deserves. The main difference between the research on discourse in NMT and SMT, apart from the general building blocks, is that the works in NMT rarely try to model discourse phenomena explicitly. On the contrary, they incorporate discourse implicitly by using sentences in the wider-document context 
via different modelling techniques, thus harnessing the inherent capability of neural models to learn discourse correlations as a by-product of optimising an NMT model. 
These studies are sometimes motivated by extensive analyses and document-level evaluation focusing on the discourse phenomena that document-level NMT improves on, as done in \cite{Jean:17, Voita:18} for anaphora, in \cite{Voita:2019} for deixis, ellipsis and lexical cohesion, and in \cite{Bawden:17} for coreference and lexical coherence/cohesion. 
Further, these works show how such models perform on automatic evaluation for MT while sometimes measuring the performance on discourse-targeted test sets. \\
\\
{\textbf{Problem Statement.}
Before going into the literature, let us first formulate the problem of document-level machine translation. Given a document $\vd$, where the set of source sentences are $\vX = \{\vx^1, \ldots, \vx^{|\vd|}\}$, the goal of document-level MT is to generate the collection of their translations $\vY = \{\vy^1,\ldots,\vy^{|\vd|}\}$ by taking into account \emph{inter-dependencies} among them imposed by the document. 

Assuming that the translations are generated left-to-right, we can decompose the probability of a document translation using the chain rule as
\begin{eqnarray}\label{eq:localdocmt}
P_{\vtheta_\local}(\vY|\vX) = \prod_{j=1}^{|\vd|} P_{\vtheta_{\textrm{local}}}(\vy^j|\vx^j, \vY^{<j},\vX^{-j})
\end{eqnarray}
where $j$ is the index of the current sentence, $\vX^{-j}$ represents all the other sentences in the source document and $\vY^{<j}$ represents the translations of the sentences preceding the $j^{th}$ sentence. This is a \emph{locally} normalised probabilistic modeling approach, which simplifies the model training (to be discussed shortly), and
 is in contrast to the \emph{globally} normalised probabilistic modelling approach \cite{andor-etal-2016-globally}, where each translation decision is conditioned on \emph{all} (past and future) translation decisions. The latter leads to greater model expressiveness to capture document-level inter-dependencies but comes at the cost of more complicated training.   
The probability of a document translation given the source document, in the globally normalised models, is
\begin{eqnarray}\label{eq:globaldocmt}
P_{\vtheta_\mglobal}(\vY|\vX) \propto \exp \sum_{j=1}^{|\vd|} \log P_{\vtheta_{\textrm{global}}}(\vy^j|\vx^j, \vY^{-j},\vX^{-j})
\end{eqnarray}
where $\vY^{-j}$ represents \emph{all} the other sentences in the target document except the $j^{th}$ sentence. Majority of the works in document-level NMT only consider the local formulation 
while a few consider the global formulation as will be evident from Tables~\ref{table:dis-nmt}, ~\ref{table:doc-nmt-extra} and~\ref{table:comparison}.\\
\\
\setlength{\tabcolsep}{2.5pt}
\begin{table}[t]
\caption{Definitions of discourse phenomena mentioned in MT literature and relevant to this survey.}
\label{table:def}
\centering
\small
  \scalebox{0.9}{
\begin{tabular}{c|p{11.75cm}}
\textbf{Discourse Phenomena} & \textbf{\hspace{5cm}{Definition}}\\
\hline
\hline
Cohesion & Cohesion is a surface property of the text and refers to the way textual units are linked together grammatically or lexically \citep{Halliday:1976}. The former,  grammatical cohesion, is based on the logical and structural content, while the latter, lexical cohesion, is based on the usage of semantically related words. \\
\hline
Pronouns & The two most popular pronoun types in the MT literature are either anaphoric
or cataphoric. Anaphoric pronouns (or pronominal anaphora) refer to someone or something mentioned previously in a text, while cataphoric pronouns refer to someone or something mentioned later in a text. The former are found to be much more common in English text \cite{Wong:2020}.\\
\hline
Coherence & Coherence refers to the underlying meaning relation between units of text and its continuity \citep{Jurafsky:2009}. It is a stronger requirement to meet than is required by cohesion, and not only embodies cohesion, but other referential components like different parts of text referring to the same entities (entity-based coherence), and relational components like connections between utterances in a discourse via coherence relations \citep{Hardmeier:2014, Smith:2018}. In other words, coherence governs whether a text is semantically meaningful overall and how easily a reader can follow it.\\
\hline
Discourse Connectives & Discourse connectives, also referred to as discourse markers or cue words, are the words that signal the existence of a specific discourse relation or discourse structure in the text. These are mostly domain-specific and may be implicit or explicit depending on the language.\\
\hline
Deixis & Deixis is reference by means of an expression whose interpretation is relative to the (usually) extra-linguistic context of the utterance, such as: who is speaking (personal deixis such as pronouns), the time or place of speaking (time and place deixis, e.g., \textit{now}, \textit{here} respectively), or the current location in the discourse (discourse deixis, e.g., \textit{this}, \textit{that}).\\
\hline
Ellipsis & Ellipsis is the omission of one or more words from a clause that is recoverable or inferable from context.\\
\hline
\end{tabular}}
\vspace{-3mm}
\end{table}

\textbf{Training.} Given a \textit{locally} normalised  document MT model, one can simply minimise the negative log-likelihood over the training set as follows,
\begin{eqnarray}\label{eq:local-train}
{\vtheta^*_{\local}}&=&\underset{\vtheta_{\local}}{\text{arg min }}\sum_{\vd\in \mathcal{D}}\sum_{j=1}^{|\vd|}{-}\log P_{\vtheta_{\local}}(\vy^j|\vx^j, \vY^{<j},\vX^{-j})\nonumber\\
&=&\underset{\vtheta_{\local}}{\text{arg min }}\sum_{\vd\in \mathcal{D}}\sum_{j=1}^{|\vd|}\sum_{n=1}^{|\vy^j|}{-}\log P_{\vtheta_{\local}}(y^j_n\mid \vy^j_{<n},\vx^j, \vY^{<j},\vX^{-j})
\end{eqnarray}
where $\mathcal{D}$ is the set of bilingual training documents. This training objective is a generalisation of the objective for sentence-level NMT given in Eq.~\ref{eq:cond}. 

Training \textit{globally} normalised document MT models can be  more challenging. This is due to the enormity of the search space over a large number of  translation variables $\vy^j$'s (i.e., the number of sentences in the document) as well as their unbounded domain (i.e., all sentences in the target language). Hence, the partition function in Eq.~\ref{eq:globaldocmt} is intractable. However, the parameters can be learned by resorting to minimising the negative \emph{pseudo-likelihood} \citep{Besag1975} as proposed in \cite{Maruf:18},
\begin{eqnarray}\label{eq:global-train}
{\vtheta^*_{\mglobal}}&=&\underset{\vtheta_{\mglobal}}{\text{arg min }}\sum_{\vd\in \mathcal{D}}\sum_{j=1}^{|\vd|}{-}\log P_{\vtheta_{\mglobal}}(\vy^j|\vx^j, \vY^{-j},\vX^{-j})\nonumber\\
&=&\underset{\vtheta_{\mglobal}}{\text{arg min }}\sum_{\vd\in \mathcal{D}}\sum_{j=1}^{|\vd|}\sum_{n=1}^{|\vy^j|}{-}\log P_{\vtheta_{\mglobal}}(y^j_n\mid \vy^j_{<n},\vx^j, \vY^{-j},\vX^{-j})
\end{eqnarray}
\noindent
\textbf{Decoding.} Instead of finding the most probable target sentence as in sentence-level NMT, for document-level NMT we need to generate the best translation for a given source document. To do so, the following optimisation problem needs to be solved:
$${\text{arg max}}_{\vY}\ P_{\vtheta}(\vY|\vX) $$
where $\vY := \{\vy^1,\ldots,\vy^{|\vd|}\}$ is the translation document, and $\vtheta$ can refer to the parameters of either locally or globally normalised models. The hardness of decoding in sentence-level MT is exacerbated for document-level MT, as the search space is enormous. 
\\

In the rest of this section, we will focus on the literature in document-level NMT which uses the wider-document context directly in the form of context sentences. While most works focus on \emph{architectures } to better model the {inter-dependencies} among the document sentences (Section~\ref{subsec:modelling}; overview in Table~\ref{table:dis-nmt}), a few focus on better \emph{training} and \emph{decoding} {strategies} (Sections~\ref{subsec:train} and~\ref{subsec:decode}; overview in Table~\ref{table:doc-nmt-extra}).\footnote{The mention of a work in a particular section is based on its main contribution.} We also present works which participated in the document-level MT shared tasks of WMT 2019 and WNGT 2019 (Section~\ref{subsec:shared-tasks}; overview in Table~\ref{table:comparison}).



\subsection{Modelling document-level context}\label{subsec:modelling}

The relevant works in document-level NMT that model the inter-dependencies among the sentences in a document can be broadly classified along two dimensions: (i) whether they use the local or the global formulation of document-level MT defined previously, and (ii) where the context comes from, that is, the source-side or both source and target-side contexts.\footnote{None of the works that model document-level context in NMT do it for only the target-side context, as it is natural to leverage the existing source-side  context.} This section has thus been divided into the four combinations that originate from the aforementioned dimensions. That is, the relevant works use either the local source-only context, the local source and target context, the global source-only context or the global source and target context.

\subsubsection{Local source-only context}
Broadly speaking, this approach is based on the following probabilistic formulation, 
\begin{eqnarray}\nonumber
P^{\src}_{\local}(\vY|\vX) = \prod_{j=1}^{|\vd|} P_{\vtheta^{\src}_{\local}}(\vy^j|\vx^j,\vX^{j\pm k}) 
\end{eqnarray}
where $k$ is the local neighbourhood of the current $j^{th}$ sentence taken to be $\{1,2,3\}$ in the existing  literature. Notice that this is not a globally normalised model, as it does not use the future context on the target-side. Further, there are works that use only the previous source sentences ($\vX^{j-k}$) and not the following sentences ($\vX^{j+k}$), as we will describe shortly. 

The first few researchers to use local source-only context for document-level NMT integrated it into the decoder. Jean et al. \cite{Jean:17} augmented the attentional RNN-based NMT architecture with an additional encoder and attention component over one previous source sentence. The context vector representation thus generated was then added as an auxiliary input to the decoder hidden state (Eq.~\ref{eq:nmtdecoder}). Through automatic evaluation and cross-lingual pronoun prediction, they found that although their approach yielded moderate improvements for a smaller training corpus, there was no improvement when the training set was much larger. Furthermore, their method suffered from an obvious limitation: the additional encoder and attention component introduced a significant amount of parameters. Hence, their method could only incorporate a few sentences in the context. 

Kuang and Xiong \cite{Kuang:2018} proposed an inter-sentence gate model to control the amount of information coming from the previous sentence while generating the translation of the current one. The main distinction of their approach from \cite{Jean:17} is that they used an inter-sentence gate to combine the two attention-based context vectors (one from the current sentence and another from the preceding one) and used the gate output as input to the decoder hidden state. They showed through experiments on Chinese$\rightarrow$English language-pair that the gate was effective in capturing cross-sentence dependencies and lexical cohesion devices like repetition.

Wang et al. \cite{Wang:17} proposed the first context-dependent NMT model to yield significant improvements over a context-agnostic sentence-based NMT model in terms of automatic evaluation. They employed a two-level hierarchical RNN to summarise the information in three previous source sentences, where the first-level RNNs were run over each of the context sentences and the second-level RNN was run over the sentence output vectors produced from the first-level ones. The final summary vector was then used to initialise the decoder, or added as an auxiliary input to the decoder state directly or after passing through a gate. Their approach showed promising results when using source-side context, noting deteriorated translation performance with target-side history. 

Voita et al. \cite{Voita:18} changed the encoder in the state-of-the-art Transformer architecture \cite{Vaswani:17} to a context-aware one. This augmented encoder comprised a source encoder and a context encoder with the first $L-1$ layers shared (see Figure~\ref{fig:transformer}). The previous source sentence served as an input to the context encoder and its output was attended to by the $L^{th}$ layer of the source encoder, and then combined with the source encoder output using a gate. The final output of the context-aware encoder was then fed into the decoder. Their experiments and pronoun-targeted analysis on  English$\rightarrow$Russian subtitles dataset 
revealed that their model implicitly learned anaphora resolution 
without the use of any specialised features. They also experimented with using the following sentence as context and found it to underperform the baseline Transformer. More recently, Wong et al. \cite{Wong:2020} found that using the future context indeed improves NMT performance both in terms of BLEU and pronoun-focused automatic evaluation. However, we believe the difference in the findings of the two works to be due to the cold-start training of the contextual NMT model in \cite{Voita:18} instead of the two-stage training approach in \cite{Wong:2020}. The latter optimises the parameters of only the sentence-level baseline in the first step while optimising the parameters of the whole network in the second step, and is shown to result in better regularisation \cite{Li:2020}.

Similar to \cite{Voita:18}, Zhang et al. \cite{Zhang:18} also used a context-aware encoder in the Transformer. However, instead of training their model from scratch, like \cite{Voita:18}, they used pre-trained embeddings from the sentence-based Transformer as input to their context encoder. In the second stage of training, they only learned the document-level parameters and did not fine-tune the sentence-level parameters of their model (following \cite{Tu:18}). They 
reported significant gains in BLEU when compared to the baseline upon experimentation with two language-pairs.

\setlength{\tabcolsep}{1.15pt}
\begin{table}[t!]
\caption{Overview of works which focus on modelling document-level context to improve NMT performance. \textit{Local document context} approaches incorporate a few context sentences in the neighbourhood of the current sentence, while \textit{global document context} approaches take the complete document into account. $s$ and $t$ denote whether the context was from the source or target-side respectively, and \textit{amount} is the maximum amount of context used in the referenced work. $^\star$$^\star$ means that the work uses discourse aspects of the document. For the \textit{Targeted Evaluation}, we only mention the phenomena which has been analysed quantitatively and ignore small-scale qualitative forms of evaluation. WSD stands for Word Sense Disambiguation.}
\label{table:dis-nmt}\vspace{-2mm}
\centering
\scalebox{0.86}
{
\begin{tabular}{c c |c|c| c|c|c|c}
\multicolumn{8}{c}{\textbf{Modelling local document context}}\\
\hline
\hline
\multicolumn{3}{c|}{\textbf{Context Type}} & \multicolumn{2}{c|}{\textbf{Approach}} & \textbf{Lang. Pair} & \textbf{Targeted} & \textbf{Reference} \\
\cline{1-5}
past & future & amount & context encoding & integration in NMT & & \textbf{Evaluation} &\\
\hline
\multirow{10}{*}{s} & \multirow{10}{*}{-} & \multirow{5}{*}{1} & \multirow{8}{*}{encoder w/attention} & \multirow{3}{*}{encoder} & En$\rightarrow$Ru & Anaphora & \cite{Voita:18} \\
\cline{6-8}
& & & & & Zh/Fr$\rightarrow$En & \multirow{2}{*}{-} & \multirow{2}{*}{\cite{Li:2020}}\\
& & & & & En$\rightarrow$De/Ru & & \\
\cline{6-8}
& & & & & Zh/Es$\rightarrow$En & - & \cite{Jiang:2019}\\
\cline{5-8}
& & & & \multirow{2}{*}{decoder} & En$\rightarrow$Fr/De & Pronouns & \cite{Jean:17}\\
\cline{6-8}
& & & & & Zh$\leftrightarrow$En & Coherence & \cite{Kuang:2018}\\
\cline{3-3}\cline{5-8}
& & \multirow{3}{*}{2} & & encoder & {En$\rightarrow$De/Tk/Ko} & {Pronouns} & {\cite{yun2020improving}}\\
\cline{5-8}
& & & & \multirow{2}{*}{encoder, decoder} & Zh/Fr$\rightarrow$En & - & \cite{Zhang:18}\\
\cline{6-8}
& & & & & Fr$\rightarrow$En & - & \cite{Wang:19}\\
\cline{3-8}
& & \multirow{2}{*}{3} & capsule network \cite{Sabour:2017} & encoder & En$\rightarrow$De & - & \cite{Yang:2019}\\
\cline{4-8}
& & & encoder & decoder & Zh$\rightarrow$En & - & \cite{Wang:17}\\
\hline
\multirow{3}{*}{s} & \multirow{3}{*}{s} & 1 & \multicolumn{2}{c|}{concatenated inputs, additional embeddings} & En$\rightarrow$De & - & \cite{Ma:2020}\\
\cline{3-8}
& & \multirow{2}{*}{2} & \multirow{2}{*}{attention} & \multirow{2}{*}{encoder} & \multirow{2}{*}{De/Pt$\leftrightarrow$En} & Pronouns & \multirow{2}{*}{\cite{Wong:2020}}\\
& & & & & & Cataphora & \\
\hline
\hline
\multirow{10}{*}{s, t} & \multirow{10}{*}{-} & \multirow{5}{*}{1} & \multicolumn{2}{c|}{concatenated inputs} & En$\rightarrow$De & - & \cite{Tiedemann:17}\\
\cline{4-8}
& & & \multirow{4}{*}{encoder w/attention} & \multirow{4}{*}{source context vector} & \multirow{3}{*}{En$\rightarrow$Fr} & Anaphora & \multirow{3}{*}{\cite{Bawden:17}}\\
& & & & & & Cohesion & \\
& & & & & & Coherence & \\
\cline{6-8}
& & & & & De/Zh/Ja$\leftrightarrow$En & - & \cite{Yamagishi:19}\\
\cline{3-8}
& & \multirow{5}{*}{3} & \multirow{3}{*}{attention} & \multirow{5}{*}{encoder, decoder} & \multirow{3}{*}{Zh/Es$\rightarrow$En} & Pronouns & \multirow{3}{*}{\cite{Miculicich:18}}\\
& & & & & & Cohesion & \\
& & & & & & Coherence & \\
\cline{4-4}\cline{6-8}
& & & \multirow{2}{*}{{encoder w/attention}} & & \multirow{2}{*}{{En$\rightarrow$Ru}} & {Deixis, Ellipsis} & \multirow{2}{*}{{\cite{xu2020efficient}}}\\
& & & & & & {Lexical cohesion} & \\
\cline{3-8}
& & \multirow{2}{*}{variable} & \multirow{2}{*}{cache} & \multirow{2}{*}{decoder} & \multirow{2}{*}{Zh$\rightarrow$En} & - & \cite{Tu:18}\\
\cline{7-8}
& & & & & & Coherence & \cite{Kuang:18} \\
\hline
\multirow{3}{*}{s, t} & \multirow{3}{*}{s} & 3 & \multicolumn{2}{c|}{\multirow{2}{*}{concatenated inputs}} & En$\rightarrow$It & - & \cite{Agrawal:18}\\
\cline{3-3}\cline{6-8}
& & {variable} & \multicolumn{2}{c|}{} & En$\leftrightarrow$De & Pronouns & \cite{Scherrer:2019}\\
\cline{3-8}
& & \multirow{2}{*}{{20}} & \multirow{2}{*}{{relative attention \cite{Shaw:2018}}} & \multirow{2}{*}{{encoder, decoder}} & \multirow{2}{*}{{Zh$\rightarrow$En, En$\rightarrow$De}} & {Deixis, Ellipsis} & \multirow{2}{*}{{\cite{Zheng:2020}}}\\
& & & & & & {Lexical cohesion} & \\
\hline
\\
\multicolumn{8}{c}{\textbf{Modelling global document context}}\\
\hline
\hline
\multicolumn{3}{c|}{\textbf{Context Type}} & \multicolumn{2}{c|}{\textbf{Approach}} & \textbf{Lang. Pair} & \textbf{Targeted} & \textbf{Reference} \\
\cline{1-5}
past & future & amount & context encoding & integration in NMT & & \textbf{Evaluation} & \\
\hline
s & - & 3$^\star$$^\star$ & encoder & encoder & En$\rightarrow$De & - & \cite{Chen:2020}\\
\hline
\multirow{3}{*}{s} & \multirow{3}{*}{s} & \multirow{2}{*}{$\star \star$} & \multicolumn{2}{c|}{\multirow{2}{*}{augmented input}} & De$\rightarrow$En/Fr & WSD & \cite{Rios:2017}\\
\cline{6-8}
& & & \multicolumn{2}{c|}{} & En$\leftrightarrow$Fr, En$\rightarrow$De & - & \cite{Mace:19}\\
\cline{3-8}
& & all & encoder & encoder, decoder & Zh/De$\rightarrow$En & Pronouns & \cite{Tan:2019}\\
\hline
\hline
s, t & - & all & encoder w/attention & decoder & Fr/De/Et/Ru$\leftrightarrow$En & - & \cite{Maruf2018}\\
\hline
\multirow{2}{*}{s, t} & \multirow{2}{*}{s, t} & \multirow{2}{*}{all} & attention & encoder, decoder & En$\rightarrow$De & Pronouns & \cite{Maruf:2019}\\
\cline{4-8}
& & & encoder w/attention & decoder, output & Fr/De/Et$\rightarrow$En & - & \cite{Maruf:18}\\
\hline
\end{tabular}
}
\vspace{-2mm}
\end{table}

The works we have mentioned so far have focused on using local contextual information to improve performance of NMT in particular. Wang et al. \cite{Wang:19} presented a more general framework for the context-aware setting, where a model is given both the source and context containing relevant information and is required to produce the corresponding output. They used document-level MT as one of the use-cases for their model. They explored three different ways of combining the source and context information in the decoder: (i) concatenating the source and context encoder outputs, (ii) adding an extra context attention sub-layer in the decoder, and (iii) interleaving the attention sub-layers by replacing the source attention sub-layer with the context attention sub-layer in the middle layers of the decoder (see Figure~\ref{fig:transformer}). Since their model encoded the source and context separately, they also presented a data augmentation technique in which they randomly removed the context information so that the model learnt to generate the target given only the source or predict the context given the source. The latter allowed the encoder to learn which parts of the context are relevant to the source sentence. They also employed a focused context attention to encourage better utilisation of the long and noisy context. They showed that their approach yields improvements over both the Transformer and a context-dependent NMT model \cite{Zhang:18} for the French$\rightarrow$English translation task on the TED Talks corpus.

{More recently, Li et al. \cite{Li:2020} conducted a study to investigate how much of the multi-encoder model improvements, as in \cite{Voita:18, Zhang:18}, come from leveraging the context. Experimentation on four small-scale datasets showed that the separate context encoder actually acts as a noise generator providing richer training signals to the NMT model, possibly alleviating overfitting inherent in such experimental settings. Further, they achieved similar improvements in BLEU even when no context was used at the decoding stage. Comparable improvements were achieved when the model was trained with random context or if Gaussian noise was added to the encoder output instead of the context. Experiment on a large-scale Chinese$\rightarrow$English corpus showed the effectiveness of using Gaussian noise while training such models. Hence, they argued for building strong sentence-level baselines with carefully regularised methods, especially when experimenting with small datasets.}

{
Jiang et al. \cite{Jiang:2019} proposed to use an associated memory network to capture the local source context for NMT. They passed each of the context sentences through an RNN encoder followed by a multi-head self-attention layer. This context representation was then combined with that of the current source sentence via an inter-sentence attention. The inter-sentence attention for each of the context sentences was then averaged and passed through a context encoder. The output of the context encoder and the standard encoder of the Transformer was further passed through a gating function. In practice, they only used one previous source sentence as context and found more number of context sentences to yield diminishing improvements. }

{Yang et al. \cite{Yang:2019} proposed to use a query-guided capsule network (QCN) \cite{Sabour:2017} to model the relationship between the words in the context and distinguish their roles with respect to the current source sentence, which acts as a query vector. Similar to \cite{Zhang:18}, they introduced a multi-head attention sub-layer in the encoder to attend to the output of the QCN. Further, they introduced a regularisation layer, which combines the source and target sentences into an identical semantic space, and computed a Pearson Correlation Coefficient loss term during the training. Experiments on English$\rightarrow$German data with the three modifications resulted in improved performance in terms of automatic evaluation on two out of the three datasets.}

Yun et al. \cite{yun2020improving} extend the idea of hierarchical context encoder introduced in \cite{Wang:17} to the Transformer architecture. However, after encoding a source context sentence using a Transformer-based context encoder, they convert it to a sentence embedding vector with a self-attentive weighted sum module. This is then concatenated over all context sentences before feeding into a second sentence-level encoder, thus resulting in a representation with correlative information from the context sentences. The final representation from the hierarchical context encoder is then combined with that from the current sentence by a gated sum in the last layer of the source encoder to generate the final encoder output similar to \cite{Voita:18}. For English$\rightarrow$Korean subtitle translation task, they demonstrated that their model closes the gap in terms of BLEU score between the sentences with helpful contexts and the sentences with unhelpful one.

{Instead of concatenating the source and context encoder outputs as in \cite{Wang:19}, Ma et al. \cite{Ma:2020} use concatenation of the context and source embeddings as inputs to their NMT model. In addition to the standard word embedding matrix, they introduced a segment embedding matrix. The word and segment embeddings (of the concatenated source and context sentences) are then added and input to the first layer of the encoder, while the remaining layers still use the current source sentence as input. This is  to allow the higher-level layers to focus more on the current source sentence and allow the context sentence to act only as supplemental semantics. Their experiments yielded comparable or slightly improved performance in BLEU over other context-aware baselines. However, they reported much better improvements when fine-tuning BERT \cite{devlin2019bert} with their approach.}

\subsubsection{Local source and target context.}
The works that fall under this category use a few previous target sentences as context in addition to the local neighbourhood of the current source sentence, thus simplifying the local formulation defined in Eq.~\ref{eq:localdocmt}. More precisely, the probability of a document translation given the source document when using the local source and target context is defined as:
\begin{eqnarray}\nonumber
P_{local}(\vY|\vX) = \prod_{j=1}^{|\vd|} P_{\vtheta_{local}}(\vy^j|\vx^j,\vY^{j-k},\vX^{j\pm k}) \hspace{1cm} k \in \{1,2,3\}
\end{eqnarray}

Tiedemann and Scherrer \cite{Tiedemann:17} were among the first researchers to experiment with the use of target context in addition to that from the source. They conducted a pilot study in which they extended the translation units without changing the underlying RNN-based NMT model. This was done in two ways: (i) extending the source sentence to include a single previous sentence, and (ii) extending both source and target sentences to include previous sentence in the corresponding context. They reported marginal improvements in terms of BLEU for German$\rightarrow$English subtitle translation, but through further analysis and manual evaluation found output examples in which referential expressions across sentence boundaries were handled properly.

Similar to \cite{Tiedemann:17}, Agrawal et al. \cite{Agrawal:18} extended the idea of concatenating translation units, but with the Transformer \cite{Vaswani:17} as base model. For the source, they experimented with up to three previous and one next sentence, while for the target, they used up to two previous sentences as the context, that is, they generated the previous and current target sentence together. They also used an RNN-based version for comparison with their models and found that, when using RNNs, concatenation underperformed multi-encoder models (similar to the ones used in \cite{Bawden:17}). They attributed this to the RNN's inherent problem of not being able to accommodate long-range dependencies in a sequence. For the Transformer, they found that the next source sentence did help in improving NMT performance, while using a large number of previous target sentences deteriorated performance due to error propagation. They concluded that the Transformer's ability to capture long-range dependencies via self-attention enabled a simple technique like concatenation of context sentences to outperform its counterpart and multi-encoder approaches with RNNs. 

Most recently, Scherrer et al. \cite{Scherrer:2019} have investigated the performance of concatenation-based context-aware NMT models in terms of different aspects of the discourse. They considered two popular datasets, the OpenSubtitles2016 \cite{Lison:16} corpus and a subset of the WMT19 corpus for the news translation task.\footnote{\url{http://www.statmt.org/wmt19/translation-task.html}} The experimental configurations for the concatenation setups were inspired from those used in \cite{Agrawal:18, junczys2019microsoft}. To test the general performance of the document-level systems, they evaluated the systems with consistent (natural order of context sentences) and artificially scrambled context (random or no context). They found that using the scrambled context deteriorated performance for the Subtitles data but not as much for the WMT data. They attributed this to the difference in the number of sentences taken as context (corresponding to a fixed number of tokens), whereby the Subtitles dataset had shorter sentences and thus much larger context than that in WMT.

Bawden et  al. \cite{Bawden:17} used multi-encoder NMT models that exploit context from the previous source sentence by combining the information from the context and current source sentence through concatenation, gating or hierarchical attention. They also experimented with an approach that decoded the previous and current target sentence together using multiple encoders on the source-side. Although they reported deteriorated BLEU scores when using the target-side context, they did highlight its importance through evaluation on test sets for coreference, coherence and cohesion.

A recent work by Yamagishi and Komachi \cite{Yamagishi:19} 
investigates models which encode source and target-side contexts via separate encoders and the ones which utilise the hidden states obtained from a pre-trained NMT baseline. The latter is referred to as weight sharing and has also been employed in \cite{Maruf:18, Miculicich:18, Maruf:2019}. This work reports that models using weight sharing almost always outperform the ones using separate context encoders for a variety of language-pairs and hypothesise this to be due to better regularisation. Further, they reached the same conclusion as \cite{Maruf:18, Miculicich:18, Maruf:2019}, that is, using target-side context is as important as the source-side context.

Miculicich et al. \cite{Miculicich:18} used three previous sentences as context by employing a hierarchical attention network (HAN) \cite{Yang:16} having two levels of abstraction: the word-level abstraction allowing to focus on words in previous sentences, and the sentence-level abstraction allowing access to relevant sentences in the context for each query word. They combined the contextual information with that from the current sentence using a gate, and used it during encoding and/or decoding a word based on if it was taken from previous source sentences and/or previously decoded target sentences respectively. Their experiments on three datasets demonstrated significant improvements in terms of BLEU. They further evaluated their model based on noun and pronoun translation, and lexical cohesion and coherence; however, they did not report whether the gains achieved for the discourse evaluation were statistically significant with respect to the Transformer baseline or not.

There have also been two approaches that use a cache to store relevant information from a document and then use this to improve the translation quality \citep{Tu:18, Kuang:18}. The first of these \cite{Tu:18} uses a continuous cache to store recent hidden representations from the bilingual context, that is the key is designed to help match the query (current context vector computed via attention) to the source-side context, while the value is designed to help find the relevant target-side information to generate the next target word. The final context vector from the cache is then combined with the decoder hidden state via a gating mechanism. The cache has a finite length and is updated after generating a complete translation sentence. They show the effectiveness of their approach on multi-domain Chinese$\rightarrow$English datasets with negligible impact on the computational cost. The second approach \cite{Kuang:18} uses dynamic and topic caches (similar to the ones in \cite{Gong:11}) to store target words from the preceding sentence translations and a set of target-side topical words semantically related to the source document, respectively. As opposed to the cache in \cite{Tu:18}, the dynamic cache  in \cite{Kuang:18} follows a first-in, first-out scheme and is updated after generating each target word. At each decoding step, the target words in the final cache are scored and a gating mechanism is used to combine the score from the cache and the one produced by the NMT model. Their experimental results on the NIST Chinese$\rightarrow$English translation task reveal that the cache-based neural model achieves consistent and significant improvements in terms of general translation quality.

Xu et al. \cite{xu2020efficient} leverage the source and target contexts by adding source-aware context encoding layer on top of the source-context encoder, context-aware source encoding layer on top of the Transformer encoder, and a context-aware target decoding layer on top of the Transformer decoder. Their main contribution is in terms of their source-context encoding, where instead of reusing the pre-trained representations from the last encoder layer (as done in \cite{Miculicich:18, Maruf:19, Maruf:2019}), they aggregate the outputs from all encoder layers by employing a layer-wise weighted combination before inputting to a self-attention sub-layer. This is then followed by a cross-attention sub-layer to attend to the current source sentence. Similar cross-attention sub-layers are used in the context-aware source encoder (to attend to the source context) and the context-aware target decoder (to attend to the context-aware source, source-aware context and the target context). This approach outperforms \cite{Zhang:18,Voita:2019} on the English$\rightarrow$Russian subtitle dataset in terms of BLEU and discourse-targeted evaluation.

Most of the approaches mentioned so far for document-level NMT combine the pre-computed higher-level abstractions of the context with the representations of the current sentence. On the contrary, Zheng et al. \cite{Zheng:2020} construct the global source context on the fly by having a segment-level relative attention on top of the Transformer encoder layers. For each sentence in the source document, the segment-level relative attention \cite{Shaw:2018} captures the inter-sentence context by taking into account the relative sentence distance. This is followed by a gated fusion to combine the intra-sentence and inter-sentence context. They also capture the target-side history by using the Transformer-XL decoder \cite{Dai:2019} which caches and reuses the hidden states computed for the previous sentence translation while generating the current one. Unlike previous work, they show that increasing the context beyond the neighbouring sentences does have a positive effect on BLEU for the Chinese$\rightarrow$English TED corpus. Further, upon visualising the segment-level relative attention, they observe that the context from the following source sentence significantly outweighs that from the previous one, thus corroborating the findings of \cite{Wong:2020}.
\subsubsection{Global source-only context.}
The works that fall under this category simplify the global formulation defined in Eq.~\ref{eq:globaldocmt}, that is, they only use discourse aspects of the source document ($\vX^{-j}$) as context and do not consider the target-side context ($\vY^{-j}$). Among these, Rios et al. \cite{Rios:2017} focused on the problem of word sense disambiguation (WSD) in NMT. One of the methods they employed to address this was to input lexical chains of semantically similar words in a document as features to the NMT model. The lexical chains were detected via learnt sense embeddings. Although this methodology did not yield substantial improvements over the baseline on a generic test set, they reported some improvement in terms of accuracy over a targeted test set that they introduced. This test set did not include any document-level context, but they found evidence that even humans were unable to resolve some of the ambiguities without access to the wider document-context.

The recent work by Mac{\'e} and Servan \cite{Mace:19} accounted for the global source context information by augmenting the source sentence with a document tag and replacing it with a document-level embedding during training. The document-level embedding was the average of the word embeddings (of words in the document) learnt while training the sentence-level model. Furthermore, the word embeddings were fixed while training the document-level model to maintain the relation between the word and the document embeddings. This minor change in the encoder input yielded promising results for both translation directions of the English-French language-pair, even though it did not yield significant improvements for English$\rightarrow$German in two out of the three test sets.

In addition to context sentences, Chen et al. \cite{Chen:2020} used the discourse structure based on Rhetorical Structure Theory (RST) \cite{Mann:88}. They retrieved the discourse tree of an input document and used the discourse path from the root to the corresponding leaf node to represent the discourse information of a word. The sequence representing the discourse path of each word was input to a Transformer-based encoder, and an average of its hidden states was used as the word discourse embedding. This discourse embedding was then added to the word embedding and input to the HAN encoder \cite{Miculicich:18}, which used three previous sentences as context.\footnote{Although this work explicitly takes only three previous source sentences as context, we consider the context to be global because each word is encoded with its discourse tree.} Thus, the role of the discourse embedding was not only to enrich the input of the encoder but also to guide it to attend to only the discourse-relevant context. The improvements in NMT performance reported in this work demonstrate the use of discourse structure in NMT to be a promising direction for future exploration.

Tan et al. \cite{Tan:2019} modelled the source-side global document context using hierarchical encoders similar to the ones in \cite{Wang:17}. Each sentence was input to a sentence-level encoder, followed by a self-attention sub-layer. The outputs of this layer were then summed over the sentences in the document and input to a document-level encoder. To assign the global context information to each word in the source sentence, the global document context from the document encoder was passed through another source-context attention sub-layer, and then added back to the word embeddings of the source sentence via a residual connection \cite{He:16}. They also introduced a document-decoder-attention sub-layer in the decoder parallel to the source-context attention sub-layer and changed the input of the feed-forward sub-layer to be the sum of the outputs of these two attention sub-layers. They tested their global context approach with both RNN and Transformer-based NMT models, and also used a large-scale sentence-parallel corpus to pre-train their models (similar to \cite{Zhang:18, Voita:2019}). The latter approach yielded greater gains in BLEU in comparison to the small-scale experiment.
\subsubsection{Global source and target context.}
To the best of our knowledge, Maruf and Haffari \cite{Maruf:18} were the first to present a document-level NMT model that successfully captured the global source and target document context using the formulation in Eq.~\ref{eq:globaldocmt}. Their model augmented the vanilla RNN-based sentence-level NMT model with external memories to incorporate documental inter-dependencies on both source and target sides via coarse attention over the sentences in the source and target documents. They used a two-level RNN to encode the source sentences in the document before applying the attention and did not perform any additional encoding for the target sentences due to risk of error propagation. They also proposed an iterative decoding algorithm based on block coordinate descent and showed statistically significant improvements in the translation quality over the context-agnostic baseline for three language-pairs. 

This work was followed up by Maruf et al. \cite{Maruf:2019}, who presented a scalable top-down approach to hierarchical attention for context-aware NMT. They used sparse attention \cite{Martins:16} to selectively focus on relevant sentences in the document context and then attend to key words in those sentences. The document-level context representation, produced from the hierarchical attention module, was integrated into the encoder or decoder of the Transformer architecture depending on whether the context was from the source or both source and target, respectively. They performed experiments and evaluation on three English$\rightarrow$German datasets in both offline (both past and future context) and online (only past context) document MT settings and showed that their approach surpassed context-agnostic and recent context-aware  baselines in most cases. Their qualitative analysis indicated that the sparsity at sentence-level allowed their model to identify key sentences in the document context and the sparsity at word-level allowed it to focus on key words in those sentences allowing for better interpretation of their document-level NMT models. 

Dialogue translation is another practical aspect of document-level MT but is underexplored in the literature. Maruf et al. \cite{Maruf2018} investigated the challenges associated with translating multilingual multi-speaker conversations by exploring the simpler task of bilingual multispeaker conversational MT. They extracted Europarl v7 and OpenSubtitles2016 to obtain an introductory dataset for the task and used complete source and target-side histories\footnote{It may seem that \cite{Maruf2018} used the local formulation of document-level MT, but we consider it to be global given the unavailability of future context for the task of dialogue translation.} in both languages as context in their model. The base architecture comprised two separate sentence-level NMT models, one for each translation direction. The source-side history was encoded using separate Turn-RNNs and then a single source-context representation vector was computed using one of five ways. The target-context representation vector was computed using a language-specific attention component. 
The source and target-side histories were incorporated in the base model's decoder separately or simultaneously to improve NMT performance. Their experiments on both public and real-world customer-service chat data \cite{Maruf19} demonstrate the significance of leveraging the bilingual conversation history in such scenarios in terms of BLEU and manual evaluation. 

\subsection{Learning with document-level context}\label{subsec:train}

All the works mentioned in Section~\ref{subsec:modelling} have proposed neural architectures which employ structural or input modifications in the base sentence-level NMT model to incorporate context information. 
For learning the parameters of their local or global document-context NMT model, these works follow the training objectives outlined for the locally or globally normalised models in Eqs.~\ref{eq:local-train} and~\ref{eq:global-train}, respectively. However, it is possible that the aforementioned objectives (or the learning process itself) may require some adjustments to accommodate for the additional context and make the learning process more efficient. 
{This includes works that induce context regularisation \cite{Jean:2019} or document-level reward functions \cite{Saunders:2020} in the training objective, modify the learning process to make the model reliant on the context \cite{Curriculum2019}, or modify the training objective to make better use of the limited training data \cite{Yu-tacl:2020} (Section~\ref{subsec:train-modify}). There is also a recent trend to utilise contextualised word embeddings to provide the document-level training process a warm-start (Section~\ref{subsec:train-embed}).}


\setlength{\tabcolsep}{1.25pt}
\begin{table}[t!]
\caption{Overview of works which focus on learning or decoding with document-level context. 
$s$ and $t$ denote whether the context was from the source or target-side respectively, and \textit{amount} is the maximum amount of context used in the referenced work. $^\star$ means that the work does not have a notion of past and future, and instead uses a sequence of sentences (fixed number of sentences or tokens) during training and/or decoding.}
\label{table:doc-nmt-extra}\vspace{-2mm}
\centering
\small
{
\begin{tabular}{c c |c|c|c|c|c}
\multicolumn{7}{c}{\textbf{Learning with document-level context}}\\
\hline
\hline
\multicolumn{3}{c|}{\textbf{Context Type}} & {\textbf{Approach}} & \textbf{Lang. Pair} & \textbf{Targeted} & \textbf{Reference} \\
\cline{1-3}
past & future & amount & & & \textbf{Evaluation} & \\
\hline
\multirow{3}{*}{s} & \multirow{3}{*}{-} & \multirow{2}{*}{1} & {learning w/context regularisation} & En$\rightarrow$Ru & - & \cite{Jean:2019}\\
\cline{4-7}
& & & {learning w/oracles} & En$\rightarrow$De & Anaphora & \cite{Curriculum2019}\\
\cline{3-7}
& & variable & \multirow{2}{*}{learning contextualised embeddings} & Zh/Fr/Es$\rightarrow$En & - & \cite{li2019pretrained}\\
\cline{1-3}\cline{5-7}
{s} & {s} & {1} & & Zh$\rightarrow$En, En$\rightarrow$De & {-} & {\cite{zhang2020learning}}\\
\hline
t & t & $\star$ & document-level minimum risk training & En$\rightarrow$De & - & \cite{Saunders:2020}\\
\hline
\multirow{2}{*}{s, t} & \multirow{2}{*}{s, t} & {$\star$} & {learning contextualised embeddings} & Zh$\rightarrow$En, En$\rightarrow$De & {-} & {\cite{liu2020multilingual}}\\
\cline{3-7}
& & all & Bayes' rule decomposition & Zh$\rightarrow$En & - & \cite{Yu-tacl:2020}\\
\hline
\\
\multicolumn{7}{c}{\textbf{Decoding with document-level context}}\\
\hline
\hline
\multicolumn{3}{c|}{\textbf{Context Type}} & {\textbf{Approach}} & \textbf{Lang. Pair} & \textbf{Targeted} & \textbf{Reference} \\
\cline{1-3}
past & future & amount & & & \textbf{Evaluation} & \\
\hline
\multirow{2}{*}{t} & \multirow{2}{*}{-} & \multirow{2}{*}{variable} & shallow fusion with semantic space LM & En$\rightarrow$Es & {-} & \cite{Garcia:2019}\\
\cline{4-7}
& & & {self-training} & {Zh$\rightarrow$En, En$\rightarrow$Ru} & {-} & {\cite{Mansimov:2020}}\\
\hline
\multirow{3}{*}{t} & \multirow{3}{*}{t} & 16 & second-pass decoding & Zh$\rightarrow$En & Coherence & \cite{Xiong:19}\\
\cline{3-7}
& & \multirow{2}{*}{$\star$} & \multirow{2}{*}{context-dependent post-editing} & \multirow{2}{*}{En$\rightarrow$Ru} & Deixis, Ellipsis & \multirow{2}{*}{\cite{Voita:19}}\\
& & & & & Lexical cohesion & \\
\hline
\multirow{2}{*}{s, t} & \multirow{2}{*}{-} & \multirow{2}{*}{3} & \multirow{2}{*}{second-pass decoding} & \multirow{2}{*}{En$\rightarrow$Ru} & Deixis, Ellipsis & \multirow{2}{*}{\cite{Voita:2019}}\\
& & & & & Lexical cohesion & \\
\hline
\end{tabular}
}
\end{table}

\subsubsection{Modifying the training strategy for document-level NMT}\label{subsec:train-modify}
Jean and Cho \cite{Jean:2019} designed a regularisation term to encourage an NMT model to exploit the additional context in a useful way 
. This regularisation term was applied at the token, sentence and corpus levels, and was based on pair-wise ranking loss, that is, it helped assign a higher log-probability to a translation paired with the correct context than to the translation paired with an incorrect one. They employed their proposed approach to train a context-aware NMT model \cite{Voita:18}, and showed that the model became more sensitive to the additional context, in addition to outperforming the context-agnostic Transformer baseline in terms of empirical evaluation (BLEU).

Recently, Saunders et al. \cite{Saunders:2020} introduced a document-level reward function in the sequence-level minimum risk training (MRT) objective for NMT \cite{Shen:2016}. They treated the batches of sentence-pairs as a pseudo-document and experimented with both random sentences and true document-context within a batch. Further, they approximated the non-differentiable gradient of the document-level metric with Monte Carlo sampling following \cite{Shannon:2017}. They showed improvements for the English$\rightarrow$German translation task when using document-level BLEU and TER \cite{Snover:06}.\textsuperscript{\ref{note:doc-metric}}

Stojanovski and Fraser \cite{Curriculum2019} introduced a curriculum learning approach \cite{Bengio:2009} that leveraged oracle information to promote anaphora resolution while training a context-aware NMT model. 
Initially, they included the gold-standard target pronouns (oracle) along with the previous context and source sentence to bias the model to pay attention to the context related to the oracle pronoun. Then, they gradually removed the oracle pronouns from the data to bias the model to pay more attention to only the context when running into ambiguous pronouns in the source sentence. The latter step was supposed to lead to superior anaphora resolution. Their scheme was unable to beat a context-aware NMT model fine-tuned with only the context in terms of both pronoun translation and overall translation quality when using a higher learning rate. However, with a lower learning rate and 25\% initial oracle samples, their approach was effective but still lagged behind the context-aware NMT model trained with higher learning rate. Their approach can be extended to other discourse phenomena, provided useful oracles are easily available. 

{Yu et al. \cite{Yu-tacl:2020} addressed the problem of lack of availability of parallel documents for document-level MT by changing the training objective to the ``noisy channel'' decomposition of the Bayes' rule, $P(\vy)\times P(\vx|\vy)$.} 
Here, $P(\vy)$ is the language model which provides unconditional probability estimates of the target document, and can be learnt from monolingual data; while $P(\vx|\vy)$ is the reverse translation probability, and requires parallel documents for its estimation. They further assumed that the sentences are translated independently at training time, allowing them to have a generative model of parallel documents, which generates source sentences in a left-to-right fashion. However, at decoding time, because of conditioning on $\vx$, inter-dependencies among the target sentences and between the target and source sentences are created \cite{Shachter:98} as is required by the definition of document-level MT. To solve the computationally complex search problem of decoding, they used an auxiliary proposal model that approximated the posterior distribution using a direct model so as to guide the search to promising parts of the output space. Their experiments showed reasonable improvements in BLEU on the Chinese$\rightarrow$English translation task.

\subsubsection{Utilising contextualised word embeddings for document-level NMT}\label{subsec:train-embed}
With the massive amounts of monolingual data and computational resources available nowadays, a new trend in natural language processing (NLP) is to learn contextualised word embeddings. Even more so, the state-of-the-art performance of such approaches on a variety of NLP tasks has added to their popularity \cite{peters2018deep, devlin2019bert}. Along similar lines for NMT, Zhang et al. \cite{zhang2020learning} proposed the idea of learning ``contextualised” embeddings of the source sentence by enforcing the NMT model to predict the local source context (in addition to the target sentence). The contextualised sentence embeddings were then integrated into the NMT model in a fine-tuning step. They proposed two methods for learning and integrating the embeddings in their Transformer-based NMT model. The first method jointly optimised the parameters for the three prediction tasks (predicting the previous and following source sentence, and the current target sentence) using parallel documents. The second method used monolingual data to optimise the parameters of a model that predicted the previous and following source sentence from the current sentence. The encoders from this step were then fine-tuned while training the NMT model. Experiments on Chinese$\rightarrow$English and English$\rightarrow$German translation tasks showed that both methods outperformed the Transformer baseline individually and when combined.

Numerous works have utilised pre-trained contextualised language models like BERT \cite{devlin2019bert} and ELMo \cite{peters2018deep} directly for improving the quality of NMT \cite{conneau2019cross,song2019mass,clinchant2019use,edunov2019pre}. However, only a few have demonstrated their systems on the document-level NMT task. Li et al. \cite{li2019pretrained} initialised the Transformer encoder with the pre-trained BERT model and also modified the encoder to control the influence of larger context. This included (i) reverse position embeddings (assign position embeddings first to the source input and then to the context) to prevent the length of the context to effect the position embeddings of the source input, and (ii) context masks in the decoder so that it learns appropriate attention weights for the current source sentence. Further, to avoid over-fitting the model on the small training corpus, they introduced the masked language model (MLM) objective on the encoder-side (same as BERT) as a term in their training objective. Their experiments on IWSLT datasets for three language-pairs showed that their system achieved the best BLEU scores among related work using the same datasets.
 
More recently, Liu et al. \cite{liu2020multilingual} extended BART, a sequence-to-sequence denoising auto-encoder comprising a bidirectional encoder (BERT \cite{devlin2019bert}) and an auto-regressive decoder (GPT \cite{Radford:2018}), by pre-training over large-scale monolingual corpora across various languages. 
For document-level MT, they followed the pre-training process introduced in the original work \cite{lewis2019bart} and extended it to documents. That is, they corrupted the input documents (of up to 512 tokens and belonging to diverse language families) by masking phrases and permuting sentences, and trained a single Transformer model to recover the original monolingual document segments. Using document fragments allowed the model to learn long-range dependencies between sentences. The main distinguishing factor of this model (mBART) from other pre-training approaches for NMT \cite{conneau2019cross,song2019mass} 
is that it pre-trains a complete auto-regressive sequence-to-sequence model end-to-end, 
once for all languages. Following this, it can be fine-tuned for any of the language-pairs in either supervised or unsupervised setting, without any task-specific or language-specific modifications or initialisation schemes. 
The results of mBART on two language-pairs in the supervised setting (feeding the source language into the encoder and decoding the target language) showed that this document-level pre-training scheme for NMT outperforms its counterpart with sentence-level pre-training. However, randomly initialised document-level NMT models (without pre-training) performed much worse than their sentence-level counterparts for both datasets. This large gap suggests that pre-training is a crucial step and a promising strategy for improving document-level NMT performance. 

\subsection{Decoding with document-level context}\label{subsec:decode}

{The approaches that we presented in Section~\ref{subsec:train} have mostly focused on introducing context during training time without any major architectural modifications to the underlying NMT model. 
They employed, however, the standard decoding scheme that followed naturally from their training process and did not account for the enormous output space (except \cite{Yu-tacl:2020}), which is inherent when generating a document translation. In this section, we present approaches that have tried to improve upon the translations obtained from a sentence-level NMT baseline by incorporating document-level context at decoding time.}

The first of these is the method proposed by Garcia et al. \cite{Garcia:2019}, who 
combined a semantic space language model (SSLM - similar to the one used in \cite{Hardmeier:12} for SMT) with a sentence-level NMT model via shallow fusion during beam-search decoding. The SSLM was supposed to promote translation choices that were semantically similar to the target context by computing a cosine similarity score between the candidate word and the non-content words preceding it in the translation. The number of the preceding words was taken to be 30 allowing for cross-sentence context. Although their approach yielded statistically significant BLEU improvements over the baseline for English$\rightarrow$Spanish translation task, the results obtained from an oracle suggest that there is still a wide margin for improvement for such fused approaches.

Similar to the two-pass approaches in SMT \cite{Xiao:11, Garcia:14, Martinez:15, Garcia:17}, there have been a few works in NMT which have focused on using context to improve upon the translations generated by a context-agnostic NMT system in a two-step process. Such approaches focus on improving document-level aspects of a text, such as, coherency and cohesion. The two-pass decoder approach by Xiong et al. \cite{Xiong:19} encourages coherence in NMT. In the first pass, they generate locally coherent preliminary translations for each sentence using the Transformer architecture. In the second step, their decoder refines the initial translations with the aid of a reward teacher \citep{Bosselut:18}, which promotes coherent translations by minimising the similarity between a sequence encoded in its forward and reverse direction. They reported gains in sentence and document-level BLEU and METEOR scores.\textsuperscript{\ref{note:doc-metric}}

Voita et al. \cite{Voita:2019} proposed a context-aware decoder to refine the translations obtained via the base context-agnostic model.\footnote{Maruf and Haffari \cite{Maruf:18} also refined the translations from sentence-level NMT baseline when using the target-side context.} However, while most previous works use the same data to train the model in both stages, they used a larger amount of data to train the sentence-level model and a smaller subset of the parallel data containing context sentences to train their context-aware model. They modified the decoder in the Transformer architecture by allowing the multi-head attention sub-layer to attend to the previous source context sentences in addition to the current source sentence. They also added an additional multi-head attention sub-layer which attended to both the previous target context sentences and the current target.\footnote{They did not mask the future target tokens when performing multi-head context attention unlike previous work \cite{Maruf:18, Maruf:2019}.} 
Although, their model performed quite well on their own targeted test sets, it yielded comparable performance with the sentence-level model in terms of BLEU. They went on to propose a context-aware model which performed automatic post-editing \cite{Voita:19} on a sequence of sentence-level translations and corrected the inconsistencies among individual translations in context of each other. The main novelty in this work is that the model was trained using only monolingual document-level data in the target language, and hence learned to map inconsistent group of sentences to their consistent counterparts. They reported significant improvements for this model in terms of BLEU, targeted contrastive evaluation of several discourse phenomenon \cite{Voita:2019} and human evaluation. 

All approaches mentioned so far have tried to enforce document-level consistency at decoding time by making changes to the decoder itself. Mansimov et al. \cite{Mansimov:2020} propose to do this by processing a document multiple times in a left-to-right fashion and self-training the sentence-level NMT model on pairs of source sentences and generated translations. To measure the upper-bound performance of this strategy, they employed an oracle which used ground-truth translations. They reported reasonable BLEU improvements in one out of the three domains they experimented with, and concluded that as long as they have a competitive sentence-level NMT model, their multi-pass self-training strategy is able to close the performance gap with the oracle.

\subsection{Shared Tasks in WMT19 and WNGT 2019}\label{subsec:shared-tasks}
Given the significant amount of work in document-level NMT in the past few years, the \textit{Fourth Conference on Machine Translation} (WMT19) \cite{wmt19} and the \textit{Third Workshop on Neural Generation and Translation} (WNGT 2019) \cite{hayashi2019findings} encouraged the use of document-level MT systems for their tasks of translating news and sports articles respectively.\footnote{{Although various shared tasks and workshops have been held for NMT in recent years, we focus only on WMT19 and WNGT 2019 in this survey as they specifically encouraged the use of document-level NMT systems. The relevant works from the \textit{Workshop on Discourse in Machine Translation} have been discussed in the previous sections.}} This opened up remarkable novelties in this domain subsuming approaches for document-level training that utilise wider document context, and also document-level evaluation. To aid this, WMT19 provided new versions of Europarl, news-commentary, and the Rapid corpus with document boundaries intact. 
They also released new versions of the monolingual Newscrawl corpus containing document boundaries for English, German and Czech. Following suit of WMT19, WNGT 2019 manually translated a portion of the RotoWire dataset\footnote{\url{https://github.com/harvardnlp/boxscore-data}}
to German. They also allowed the use of any parallel and monolingual data made available by the WMT19 English-German news task and the full RotoWire English dataset.

To evaluate the document-level MT systems, WMT19 introduced two human evaluation setups to reliably compare document-level MT systems and human performance. WNGT 2019, however, only relied on automatic evaluation based on the BLEU score.
\subsubsection{Document-level MT systems in WMT19.} 
Among the 153 submissions for the news translation task at WMT19, only a few utilised document-level context. Stojanovski and Fraser \cite{Stojanovski:2019} introduced a context-aware NMT model which modelled the local context (previous  sentence), also taking advantage of the larger context (previous ten sentences). The main architecture is similar to what they previously proposed in \cite{Curriculum2019} and includes the previous sentence explicitly. For the larger context, they created a simple document-level representation by averaging word embeddings, which were added to the source embeddings similar to the positional encodings in the Transformer. They showed that most gains over the baseline come from the addition of the implicit larger context.

\renewcommand\theadfont{\normalsize}
\begin{table}[t!]
\caption{Overview of document-level systems in WMT2019 and WNGT2019 shared tasks. $s$ and $t$ denote whether the context was from the source or target-side respectively, and \textit{amount} is the maximum amount of context used in the referenced work. $^\star$ means that the work does not have a notion of past and future and instead uses a sequence of sentences (fixed number of tokens) during training and decoding. $^\star$$^\star$ means the system incorporates discourse aspects of the document. All systems use parallel sentences and task-specific parallel documents for training. Some works also utilise \textit{synthetic parallel documents} produced by putting document boundaries on sequence of random sentences. 
    {BLEU scores for the best document-level models on the newstest2018 (WMT19) and RotoWire test (WNGT19) for systems participating in the English$\rightarrow$German translation task are reported.} None of the systems perform large-scale discourse-targeted evaluation.}
    \label{table:comparison}\vspace{-2mm}
    \centering
\scalebox{0.82}{
 \begin{tabular}{c|c||c c|c|c|c|c|c|c|c|}
    \multirow{4}{*}{{\textbf{Venue}}}& \multirow{4}{*}{\textbf{System}}& \multicolumn{3}{c}{\textbf{Context Type}}& \multicolumn{1}{|c}{\textbf{Parallel Documents}}& \multicolumn{3}{|c}{\textbf{Training Techniques}} & \multicolumn{1}{|c}{\textbf{Decoding}} & \multicolumn{1}{|c|}{{\textbf{BLEU}}}\\ 
        \cline{3-9}
         &  & \thead{past} & \thead{future} & \thead{amount} & \thead{synthetic} & \thead{back\\ translation} &
         \thead{fine-tuning} & 
         \thead{modified \\ architecture} & \textbf{w/context} & \thead{{best}\\ {model}}\\
         \cline{5-7}
         \hline
         \multirow{6}{*}{\textbf{\textit{WMT19}}} & 
         \cite{Stojanovski:2019}
         & s & - & 10 & - & \checkmark &\checkmark & \checkmark &-& {47.1}\\ 
          \cline{2-11}& 
         \cite{espana2019uds} 
         & s & s & $\star \star$ & - & - & - & - & -& {42.8}\\ 
         \cline{2-11} & 
         \cite{stahlbergcued}
         & t & - & \multirow{2}{*}{all} & - & \checkmark & \checkmark& \checkmark&-& {49.3}\\ 
         \cline{2-4}\cline{6-11} & 
         \cite{talman:2019} & s, t & s & & - & \checkmark & - & \checkmark & \checkmark & {39.1}\\
         \cline{2-11} & 
         \cite{junczys2019microsoft}&
         \multirow{4}{*}{s, t} & \multirow{4}{*}{s, t} & \multirow{3}{*}{$\star$} & \checkmark & \checkmark & \checkmark& - & \checkmark & {50.3}\\ 
         \cline{2-2}\cline{6-11} & 
         \cite{popel2019english}&
         & & & \checkmark& \checkmark &\checkmark & - & \checkmark & {-}\\ 
         \cline{1-2}\cline{6-11}
         \multirow{2}{*}{\textbf{\textit{WNGT19}} } & 
         \cite{saleh2019naver} & & & & \checkmark &\checkmark& \checkmark& - & - & {48.0}\\ 
          \cline{2-2}\cline{5-11} & 
         \cite{Maruf:19}
         & & & all & - & - & - &\checkmark &\checkmark & {41.5}\\ 
         \hline
    \end{tabular}
    }
    \vspace{-3mm}
\end{table}

Espa{\~n}a-Bonet et al. \cite{espana2019uds} enriched the document-level data by adding coreference tags in the source sentences, where the tags were obtained by running a mention-ranking model (\texttt{CoreNLP} \cite{Manning:2014}) over the source documents. Since a sentence had only a few (or no) annotated words, their approach did not yield any significant improvements, but achieved minor gains when using ensembling.

The Cambridge University Engineering Department's system \cite{stahlbergcued} relied on document-level language models (LMs) to improve the sentence-level NMT system. They modified the Transformer architecture for document-level language modelling by introducing separate attention layers for inter- and intra-sentential context. The LM was trained independently of the translation model and acted as a post-editor for the sentence-level translations when the context was available. They reported minor improvements in BLEU over strong baselines for their approach. 

Talman et al. \cite{talman:2019} experimented with concatenating the local source and target context with the current sentence in various ways, including the variation proposed in \cite{Tiedemann:17}. However, they reported deteriorated performance in terms of BLEU when comparing their approach to the sentence-level Transformer baseline. They also experimented with hierarchical approaches to document-level NMT \cite{Miculicich:18, Maruf:19} but were unable to obtain any gains over the sentence-level counterparts.

The Microsoft Translator \cite{junczys2019microsoft} introduced three systems, two of which focused on large-scale document-level NMT with 12-layer Transformer-Big systems \cite{Vaswani:17}. For the first system, they combined real document-parallel data with synthetic document-parallel data (produced by putting in document boundaries at random) and created document-level sequences of up to 1000 subword units to train deep transformer models. In addition, they used back-translated documents and further fine-tuning techniques. Their second document-level system comprised a BERT-style encoder, trained on monolingual English documents, sharing its parameters with the translation model. They also experimented with second-pass decoding and ensembling techniques to combine the sentence-level systems with the document-level ones. Based on human evaluation, it was found that their document-level systems were preferred over the sentence-level ones.

Popel et al. \cite{popel2019english} proposed document-level NMT systems for English$\rightarrow$Czech implemented in Marian \cite{marian2018} and T2T \cite{t2t2018} frameworks. 
They extracted context-augmented data from real and synthetic document-level data (created using back-translation) by creating sequences of consecutive sentences with up to 1000 tokens
. They initially trained sentence-level baselines using the same setting as \cite{popel2018cuni} and then fine-tuned their model using the context-augmented data. They experimented with different document-level decoding strategies for both frameworks. For Marian, they translated up to three consecutive sentences at once and allowed a sentence to appear at all possible positions, such that a sentence was translated at most six times. The final translation of a sentence was selected based on a ranking of the six translation setups obtained from the validation set BLEU scores. 
For T2T, each document was split into overlapping multi-sentence segments consisting pre-context (only used to improve the translation of the main content), main content and post-context (similar to pre-context). The final translation hypothesis was selected to be the ``middle” sentence in a sequence (corresponding to the main content). They were unable to achieve significant improvements over the sentence-level baseline with any framework.
\subsubsection{Document-level MT systems in WNGT 2019. }
WNGT 2019 was not as popular as WMT19 in terms of the number of submissions, having received only four for the document translation task. Out of these, only two systems incorporated document-level context. 

Naver Labs Europe's system \cite{saleh2019naver} relied heavily on transfer learning from document-level MT (high-resource) to document-level NLG (low-resource). They first trained a sentence-level MT model with WMT19, RotoWire and back-translated Newscrawl data. A domain-adapted document-level MT system was then trained via two levels of fine-tuning: (i) fine-tuning the best sentence-level model (chosen based on validation set perplexity) on real and synthetic document-level data, (ii) fine-tuning the best document-level model on document-level Rotowire plus back-translated monolingual Rotowire data (in-domain). They ranked first in all tracks of the document-translation task outperforming the baseline by at least 11.76 BLEU score.

Maruf and Haffari \cite{Maruf:19} re-used their previously proposed document-level NMT model \cite{Maruf:2019} for the shared task. 
For the first stage of training, they used sentence-level data from Europarl, Common Crawl, News Commentary v14, the Rapid corpus and the Rotowire parallel data, while for the second stage, they used document-level data from the latter three corpora. They did not perform any further fine-tuning on the Rotowire parallel dataset. For decoding, they ensembled three independent runs of all models using two ensemble decoding strategies, which involved averaging the probability distributions at the softmax level with and without averaging the context representations obtained from the individual runs. For both translation directions, their hierarchical attention approach surpassed the WNGT baseline by at least 4.48 BLEU score even without in-domain fine-tuning, showing that their approach is indeed beneficial even in a large-data setting.

The MT results of the submissions on DGT task show that pre-training with back-translated data and fine-tuning document-level MT models on document-level in-domain data leads to drastic improvements. It has also been shown that adding structured-data for this specific task does not lead to improvements over the baselines \cite{hayashi2019findings}.

\subsection{Summary}

It should be noted from the aforementioned works that document-level NMT has flourished significantly in the past three years, mostly due to the improvements in translation quality that we were unable to see with SMT techniques. However, most methods only exploit a small context beyond a single sentence, comprising neighbouring sentences (\textit{local context}), and do not incorporate context from the whole document (\textit{global context}). At this point in time, it is unclear whether the global context is indeed useful due to the lack of discourse-targeted analysis undertaken by numerous cited works (as is evident from Tables~\ref{table:dis-nmt}, ~\ref{table:doc-nmt-extra} and~\ref{table:comparison}). Nonetheless, it should be noted that global context approaches are more general and can be simplified to the use of local context, while the opposite is either not possible or involves greater effort. 

A recent study by Kim et al. \cite{Kim:2019} made a case against the usefulness of wider document-context as opposed to various works covered in this section. Firstly, they found that filtering the redundant or irrelevant words in the context does not harm the translation quality measured via BLEU in comparison to when using full context sentences. The document-level NMT models that they experimented with were also unable to outperform a baseline trained on a larger corpus, and thus they attributed the improvements gained by the current document-level NMT models to better regularisation (similar to the conclusions drawn in \cite{Li:2020}). They also point out that the document-level context seldom improves discourse aspects in the translation, for instance coreference and lexical choice, as opposed to what has been demonstrated by previous works \cite{Voita:18, Miculicich:18, Voita:2019}. They conclude that the improvements are mostly general in terms of adequacy and fluency, and these are not an interpretable way of utilising the document-level context. 

In light of this work and numerous others that we have cited in this survey, we believe that 
one study is not sufficient to invoke skepticism of the success achieved by document-level NMT. Having said that, to sustain the progress in this domain, it is also necessary that the use of document-level NMT models should be motivated by more in-depth and painstaking discourse-targeted analysis.

\section{Evaluation}\label{sec:diseval}
MT outputs are almost always evaluated using metrics like BLEU and METEOR, which use n-gram overlap between the translation and reference sentence to judge translation quality. However, these metrics do not look for specific discourse phenomena in the translation, and thus may fail when it comes to evaluating the quality of longer pieces of generated text.\footnote{\label{note:doc-metric}Some works \cite{Gong:15, Xiong:19, Saunders:2020, liu2020multilingual} extend the standard automatic MT metrics to documents by concatenating the sentences within a document into one long sentence and then applying the traditional metric to it. This type of evaluation is still limited as it does not consider discourse aspects of the text.} A recent study by L{\"a}ubli et al. \citep{Laubli:18} contrasts the evaluation of individual sentences and entire documents with the help of human raters. They found that the human raters prefer human translations over machine-generated ones when assessing adequacy and fluency of translations. Hence, as translation quality improves, there is a dire need for document-level MT evaluation methods since errors pertaining to discourse phenomena remain invisible in sentence-level evaluation. 

There has been some promising work towards proposing new evaluation metrics for specific discourse phenomena, but there is no consensus among the MT community about their usage. Most of these metrics perform evaluation based on a single reference sentence, with the exception of \cite{Smith:18}, without taking the context into account (except \cite{Jwalapuram:19}) 
 as will be described in Section~\ref{subsec:auto-eval}. There are also those that suggest using evaluation test sets (Section~\ref{subsec:test-eval}) or better yet combining them with semi-automatic evaluation schemes \citep{Guillou:18}. More recently, Stojanovski and Fraser \cite{Stojanovski:18} proposed to use oracle experiments for evaluating the effect of pronoun resolution and coherence in MT. Table~\ref{table:dis-eval} outlines the works introducing document-level MT evaluation approaches. For a more elaborate review of studies that evaluate NMT output specifically in terms of analysing its limitations when translating discourse phenomena, we direct the reader to \cite{Popescu-Belis:19}.

\begin{table}[t!]
\caption{Overview of works which introduce techniques to evaluate discourse phenomena in MT. Some of the referenced works do not consider inter-sentence context information.}
\label{table:dis-eval}\vspace{-2mm}
\centering
\small
{
\begin{tabular}{c|c|c|c}
\textbf{Evaluation Type} & \textbf{Discourse Phenomena} & \textbf{Dependency} & \textbf{Reference}\\
& & \textbf{(Resource or Language)} & \\
\hline
\hline
Automatic Metric & Pronouns & Alignments, Pronoun lists & \cite{Hardmeier:10, Miculicich:17}\\
 & & English in target (anaphoric) & \cite{Jwalapuram:19} \\
\cline{2-4}
& Lexical Cohesion & Lexical cohesion devices & \cite{Wong:2012}\\
& & Topic model, Lexical chain & \cite{Gong:15} \\
\cline{2-4}
& Discourse Connectives & Alignments, Dictionary  & \cite{Hajlaoui:13}\\
& & Discourse parser & \cite{Guzman:2014, Joty:2017, Smith:18}\\
\hline
Test Suite & Pronouns &En$\rightarrow$Fr & \cite{Guillou:16}\\
& &En$\rightarrow$Fr (anaphora) & \cite{Bawden:17}\\
& &En$\rightarrow$De (anaphora) & \cite{Mueller:18} \\
\cline{2-4}
& Cohesion  &En$\rightarrow$Fr & \cite{Bawden:17}\\
& & En$\rightarrow$Ru & \cite{Voita:2019}\\
\cline{2-4}
& Coherence  &En$\rightarrow$Fr & \cite{Bawden:17}\\
& &En$\leftrightarrow$De,  Cs$\leftrightarrow$De, En$\rightarrow$Cs& \cite{vojtvechova2019sao}\\
& &En$\rightarrow$Cs &\cite{rysova-etal-2019-test}\\
\cline{2-4}
& Conjunction & En/Fr$\rightarrow$De &\cite{popovi:2019}\\
\cline{2-4}
& Deixis, Ellipsis &En$\rightarrow$Ru& \cite{Voita:2019}\\
\cline{2-4}
& Grammatical Phenomena & En$\rightarrow$De & \cite{Sennrich:2017}\\
& & De$\rightarrow$En & \cite{avramidis2019linguistic} \\
\cline{2-4}
& Word Sense Disambiguation & De$\rightarrow$En/Fr & \cite{Rios:2017, Rios:2018}\\
& & En$\leftrightarrow$De/Fi/Lt/Ru, En$\rightarrow$Cs&\cite{raganato2019mucow}\\
\hline
\end{tabular}
}
\vspace{-2mm}
\end{table}

\subsection{Automatic Evaluation of Specific Discourse Phenomena} \label{subsec:auto-eval}
In this section, we discuss automatic metrics for evaluating specific discourse phenomena in MT. \\
\\
\textbf{Pronoun translation.} The first pronoun evaluation metric \cite{Hardmeier:10} measures the precision and recall of pronouns directly. Firstly, word alignments are produced between the source and translation output, and the source and the reference translation. For each pronoun in the source,  a clipped count -- the number of times the pronoun occurs in the translation output truncated by the number of times it occurs in the reference translation -- is computed. The final metric is then the precision, recall or F-score based on the clipped counts. 

Werlen et al. \cite{Miculicich:17} estimated the accuracy of pronoun translation (APT), that is for each source pronoun, they counted the number of times its translation is considered correct. APT first identifies triples of pronouns: (source pronoun, reference pronoun, candidate pronoun) based on word alignments, improved through heuristics. Next, the translation of a source pronoun in the MT output and the reference are compared, and the number of identical, equivalent, or different/incompatible translations in the output and reference, as well as cases where candidate translation is absent, reference translation is absent or both, are counted. Each of these cases is assigned a weight between 0 and 1 to determine the level of correctness of MT output given the reference. The weights and the counts are then used to compute the final score. 

Most recently, Jwalapuram et al. \cite{Jwalapuram:19} proposed a specialised measure for pronoun evaluation trained to distinguish a good translation from a bad one based on pairwise evaluations between two candidate translations (with or without past context). The evaluation is independent of the source language and is shown to be highly correlated with human judgments. They also presented a pronoun test suite that covers multiple source languages and various target pronouns in English. Both their test set and evaluation measure were based on actual MT system outputs. \\
\\
\textbf{Lexical cohesion.} Wong and Kit \cite{Wong:2012} extended the sentence-level evaluation metrics, like BLEU \citep{Papineni:02}, METEOR \citep{Banerjee:05} and TER (Translation Edit Rate) \citep{Snover:06}, to incorporate a feature that scores lexical cohesion. To compute the new score, they identify lexical cohesion devices via WordNet clustering \citep{fellbaum:98} and repetition via stemming, and then combine this score with the sentence-level one through a weighted average. They claimed that this new scoring feature increases the correlation of BLEU and TER with human judgments, but does not have any effect on the correlation of METEOR. Along similar lines, Gong et al. \cite{Gong:15} augmented document-level BLEU and METEOR\textsuperscript{\ref{note:doc-metric}} 
with a cohesion score and a gist consistency score using a weighted average. The cohesion score is computed based on simplified lexical chain, while the gist consistency score is based on topic model. Their hybrid metrics could obtain significant improvements for BLEU but only slight improvements for METEOR.\\
\\
\textbf{Discourse connectives.} Hajlaoui and Popescu-Belis \cite{Hajlaoui:13} proposed new automatic and semi-automatic metrics referred to as ACT (Accuracy of Connective Translation) \citep{Meyer:2012}. For each connective in the source, ACT counts one point if the translations are the same and zero otherwise based on a dictionary of possible translations and word alignments. The insertion of connectives is counted manually. The final score is the total number of counts divided by the number of source connectives. Smith and Specia \cite{Smith:18} proposed a reference-independent metric that assesses the translation output based on the source text by measuring the extent to which the discourse connectives and relations are preserved in the translation. Their metric combines bilingual word embeddings pre-trained for discourse connectives with a score reflecting the correctness of the discourse relation match. However, their metric depends on other lexical elements, such as a parser, which may miss some constituents or discourse relations.

Discourse structure has also been used for improving MT evaluation \cite{Guzman:2014, Joty:2017}. Guzman et al. \cite{Guzman:2014} developed a discourse-aware evaluation metric by first generating discourse trees for the translation output and reference using a discourse parser (lexicalised and un-lexicalised), and then computing a similarity measure between the two trees. This metric is based on the assumption that good translations would have a similar discourse structure to that of the reference. \\
\\
Guillou and Hardmeier \cite{Guillou:18} studied the performance of automatic metrics for pronouns \cite{Hardmeier:10, Miculicich:17}, on the PROTEST test suite \citep{Guillou:16} of English$\rightarrow$French translations and explored the extent to which automatic evaluation based on reference translations can provide useful information about an MT system's ability to handle pronouns. They found that such automatic evaluation can capture some linguistic patterns better than others and recommend emphasising high precision in the automatic metrics and referring negative cases to human evaluators. It has also been suggested that to take MT to another level, ``the outputs need to be evaluated not based on a single reference translation, but based on notions of fluency and of adequacy -- ideally with reference to the source text'' \citep{Smith:2017}.

\subsection{Evaluation Test Sets} \label{subsec:test-eval}

As pointed out in \cite{graham2019translationese, toral2018parity, Laubli:18}, a tie of human and machine when both are evaluated on the basis of isolated translation segments cannot serve as an indication of \textit{human parity}. The current document-level evaluations, for instance the DR+DC evaluation setup in WMT19, are not reliable due to their low statistical power (small sample size) of the rated documents \cite{graham2019translationese}. Furthermore, even the best translation systems may fall behind when it comes to handling discourse phenomena \cite{vojtvechova2019sao}. These highlighted shortcomings explain the motivation of using targeted test suites which can draw better conclusions about whether or not a machine achieves human parity, in addition to aiding a more in-depth analysis of various aspects of the translation.\\
\\
\textbf{Contrastive test sets.} A few of the discourse-targeted test sets are \textit{contrastive}, that is each instance contains a correct translation and a few incorrect ones. Models are then assessed on their ability to rank the correct translation of a sentence in the test set higher than the incorrect translation. Sennrich et al. \cite{Sennrich:2017} were the first to introduce a large contrastive test set to evaluate five types of grammatical phenomena known to be challenging for English$\rightarrow$German translation. 
Even though contextual information may be beneficial for such phenomena, this particular test suite did not contain any, due to which we will not present details of this and any similar test suites any further. Following \cite{Sennrich:2017}, Rios et al. \cite{Rios:2017, Rios:2018} also introduced a contrastive test set for word sense disambiguation, but again, it did not contain document-level context. However, the document context can be recovered as they have provided the corpora from which the test set was extracted. 

Inspired by examples from OpenSubtitles2016 \citep{Lison:16}, Bawden et al. \cite{Bawden:17} hand-crafted two contrastive test sets for evaluating anaphoric pronouns, coherence and cohesion in English$\rightarrow$French translation. All test examples were designed so as to have the particular phenomena in the current English sentence ambiguous such that its translation into French relied on the previous context sentence. Hence, these test sets required a model to leverage the previous source and target sentences in order to improve the said phenomena. M{\"u}ller et al. \cite{Mueller:18} presented a contrastive test suite to evaluate the accuracy with which NMT models translate the English pronoun \textit{it} to its German counterparts \textit{es}, \textit{sie} and \textit{er} while having access to a variable amount of previous source sentences as context. Voita et al. \cite{Voita:2019} created English$\rightarrow$Russian test sets focused on deixis, ellipsis and lexical cohesion as they found these three phenomena to be the cause of 80\% of the inconsistencies in Russian translations. They showed that their context-aware NMT model 
performed substantially well on these test sets, even though it did not show significant gains over the context-agnostic baseline (Transformer) in terms of general translation quality measured via BLEU. 
\\
\\
\textbf{Test sets with document-level information at WMT19.} WMT19 \cite{wmt19} also went beyond the rating evaluation by providing complimentary test suites. Vojt{\v e}chov{\'a} et al. \cite{vojtvechova2019sao} introduced the SAO (Supreme Audit Office of the Czech Republic) WMT19 Test Suite 
to evaluate the domain-independent performance and document-level coherence 
when testing MT models trained for the news domain on 
audit reports. They provided a tri-parallel test set consisting English, Czech and German documents in addition to documents in Polish and Slovak. Their evaluation experiment on the systems submitted for WMT19 news task showed that although the MT systems could perform quite well for audit reports in terms of automatic MT evaluation, a thorough in-domain knowledge was required for assessing aspects like semantics and domain-specific terminology which was further restricted due to the use of a single reference translation. 
They conclude that the most practically reliable solution for translation would be an interactive system supporting a domain expert who manually amends the machine terminological choices. 

Rysov{\'a} et al. \cite{rysova-etal-2019-test} introduced an English$\rightarrow$Czech test suite to assess the translation of discourse phenomena in the WMT19 systems. After manually evaluating the translations, they identified translation errors for three document-level discourse phenomena: topic-focus articulation (information structure), discourse connectives, and alternative lexicalisation of connectives. 
The evaluation results showed that WMT19 systems almost reached the reference quality on this test suite, while document-level MT systems were reported to have the most errors for the category of topic-focus articulation, showing satisfactory performance on the other two phenomena. 

The test suites introduced in \cite{avramidis2019linguistic, raganato2019mucow, popovi:2019} were also used to evaluate document-level MT systems at WMT19 but these consisted only sentence-pairs and did not rely on any extra-sentential context information. We conclude that an evaluation using test suites is feasible but has a restricted scope since it is designed for specific language-pairs and provides limited guarantees.


\section{Conclusions and Future Directions}\label{sec:conc}

{ In this survey, we have presented a comprehensive overview of the research that has incorporated the wider document-context in NMT. We started off by providing a brief introduction to the standard sentence-level NMT models and their evaluation framework (Section~\ref{sec:nmt}). After presenting the foundations, we delved deep into the literature of document-level NMT (Section~\ref{sec:disnmt}) and categorised the relevant research based on its main contributions -- modelling document-level context or learning/decoding with document-level context.
We further organised the literature on modelling document-level context for NMT based on whether it captures: (i) the local or global context,  and (ii) the source or both source and target-side context. 
Lastly, we covered the evaluation strategies employed to measure the quality of document-level MT systems (Section~\ref{sec:diseval}).
}

Despite the progress that document-level MT has seen due to the end-to-end learning framework provided by neural models, there is still a lot that needs to be done, not only in terms of better modelling of the context but also context-dependent evaluation strategies. Let us now mention a few of the possible future research directions.\\
\\
\textbf{Document-aligned Datasets.} 
While there are many popular datasets for MT, most of them consist of aligned sentence-pairs. Hence, one of the main problems faced by the community is the curation of document-parallel bilingual datasets. Particularly, some targeted discourse phenomena do not exist in the current public datasets. This problem further exacerbates when one tries to translate dialogues since existing datasets, such as movie subtitles, lack speaker annotations. 
Furthermore, the research community should pay attention to multi-domain and morphologically rich languages {(as done in \cite{liu:2020})}, which are excellent use-cases to test the actual worth of document-level MT approaches.
It is the right time  to invest efforts in creating such resources (as initiated by WMT19) so that the research process can be standardised with respect to the datasets used. 
\\
\\
\textbf{Explicit Discourse-level Linguistic Annotation.}
If the process of obtaining discourse annotations could be automated, it can affect both the development and evaluation of document NMT systems. For instance,  if we could obtain annotations of  discourse entities, it could directly impact the translation of their mentions, thus improving lexical cohesion. The translation could also be conditioned on the evolution of entities as they are introduced in the source and target text \citep{Ji:17}. We believe annotation of discourse phenomena, such as coreference or discourse markers, is critical to the advancement of the field. 
\\
\\
\textbf{Document-level MT Evaluation.} From the previous section, it is evident that there is no consensus among the MT community when it comes to the evaluation of document-level MT. Reference-based automatic evaluation metrics, like BLEU and METEOR, 
are insensitive to the underlying discourse structure of the text \citep{Laubli:18}. These are still being used to evaluate MT outputs as they have been the de-facto standard in the community for almost two decades. The proposed document-level automatic metrics (detailed in Section~\ref{sec:diseval}) have their own flaws and are not widely accepted. Moreover, comparison to a single reference translation is not a good way to evaluate translation output as it has its own shortcomings. A middle ground should be found between automatic and manual evaluation for document MT that could make the process of manual evaluation cheaper, and would still be better than the current automatic metrics at evaluating discourse phenomena. Evaluation test sets only resolve a part of the problem as they are mostly hand-engineered for specific language-pairs. To actually progress in document-level MT, we not only need models that address it, but also evaluation schemes that have the ability to correctly gauge their performance.

This survey presents a resource to highlight  different aspects of the literature in document-level MT. We hope it makes it easier for researchers to take stock of where this research field stands, and identify the accomplishments and avenues for future research to further flourish it. 



\bibliographystyle{plain}


\end{document}